\documentclass{article}

\usepackage{microtype}
\usepackage{graphicx}
\usepackage{subcaption}
\usepackage{booktabs}
\usepackage{multirow}
\usepackage{hyperref}

\usepackage[preprint]{icml2026}
\hypersetup{pdfsubject={}}
\usepackage{amsmath}\usepackage{amssymb}\usepackage{mathtools}\usepackage{amsthm}
\usepackage[capitalize,noabbrev]{cleveref}

\newif\ifvenueicml \venueicmltrue
\newif\iflongform  \longformfalse
\newenvironment{widefigure}[1][tbp]{\begin{figure*}[#1]}{\end{figure*}}
\newenvironment{widetable}{\begin{table*}}{\end{table*}}

\newcommand{\Kmodes}{K}                       
\newcommand{\modek}{k}                        
\newcommand{\paramz}{x}                       
\newcommand{\paramdim}[1]{d_{#1}}             

\newcommand{\deter}{h}
\newcommand{\stoch}{\mathbf{z}}
\newcommand{\imhorizon}{H}
\newcommand{\discount}{\gamma}
\newcommand{\Ncand}{C}                        

\newcommand{\Platform}{Platform}
\newcommand{\Goal}{Goal}
\newcommand{\HardGoal}{Hard Goal}
\newcommand{\CatchPoint}{Catch Point}
\newcommand{\HardMove}[1]{Hard Move\,($n{=}#1$)}
\newcommand{\HardMoveN}{Hard Move}           
\newcommand{\HM}[1]{HM#1}                     

\newcommand{\ours}{\textsc{HyLI}}    
\newcommand{\oursshort}{\ours{} (ours)}     
\newcommand{\dlpa}{DLPA}
\newcommand{\hyar}{HyAR}
\newcommand{\pdqn}{P-DQN}
\newcommand{\hhqn}{HHQN}
\newcommand{\hppo}{H-PPO}
\newcommand{\patdd}{PA-TD3}
\newcommand{\chdp}{CHDP}
\newcommand{\daac}{DA-AC}

\newcommand{\cme}{counterfactual mode evaluation}   
\newcommand{\Cme}{Counterfactual mode evaluation}
\newcommand{\CME}{CME}
\newcommand{\swr}{success-window replay}            
\newcommand{\SWR}{SWR}
\newcommand{\Swr}{Success-window replay}
\newcommand{\mureg}{$\mu$-regularization}           

\newcommand{\muregshort}{$\mu$-reg}
\newcommand{\epsmode}{$\varepsilon$-mode exploration} 
\newcommand{\Epsmode}{$\varepsilon$-mode exploration}
\newcommand{\epsshort}{$\varepsilon$-mode}

\newif\ifdraft \draftfalse
\usepackage{xcolor}
\usepackage{float}                            
\ifdraft
  \newcommand{\note}[1]{\textcolor{red}{[#1]}}
  
\else
  \newcommand{\note}[1]{}
  
\fi

\ifdraft
  
\else
  
\fi

\newcommand{\cNmatch}{7}
\newcommand{\cNabove}{3}

\newcommand{\cNbestMean}{5}

\newcommand{\cEffHyar}{5.0}

\newcommand{\cEffHyarNoPre}{2.4}
\newcommand{\cEffHyarNoPreMin}{1.4}
\newcommand{\cEffHyarNoPreMax}{5.8}
\newcommand{\cNfastestDlpa}{5}

\newcommand{\cNfastestOursRealtime}{7}
\newcommand{\cDlpaRatioCpu}{67}
\newcommand{\cDlpaRatioMinCpu}{40}
\newcommand{\cDlpaRatioMaxCpu}{286}
\newcommand{\cOursMinHzCpu}{640}
\newcommand{\cDlpaMaxHzCpu}{22}
\newcommand{\cOursMsCpu}{1.1--1.6}
\newcommand{\cMfRatioMinCpu}{3}
\newcommand{\cMfRatioMaxCpu}{17}
\newcommand{\cDlpaHmTenMsCpu}{1145}

\newcommand{\cDlpaMaxHzGpu}{33}

\newcommand{\cReproN}{35}
\newcommand{\cReproIn}{24}
\newcommand{\cReproLow}{7}
\newcommand{\cReproHigh}{4}

\newcommand{\cEpLenGoalMin}{3}
\newcommand{\cEpLenGoalMax}{6}
\newcommand{\cDeadFour}{1.2}
\newcommand{\cDeadFourUni}{25}
\newcommand{\cDeadSix}{0.5}
\newcommand{\cDeadSixUni}{16}
\newcommand{\cDeadEight}{0.6}
\newcommand{\cDeadEightUni}{6}

\newcommand{\cLadGBaseFail}{3}
\newcommand{\cLadGBaseIqm}{0.24}
\newcommand{\cLadGBaseN}{5}

\newcommand{\cLadGCmeDelta}{+0.09}
\newcommand{\cLadGCmeDeltaHi}{+0.16}
\newcommand{\cLadGCmeDeltaLo}{+0.00}

\newcommand{\cLadGCmeSteps}{63.5k}

\newcommand{\cLadGEpsSteps}{40k}

\newcommand{\cLadGPaBaseIqm}{0.49}

\newcommand{\cLadGPaSwrDelta}{$-$0.10}

\newcommand{\cLadGPaSwrIqm}{0.57}

\newcommand{\cLadGPaSwrSwrDelta}{+0.08}

\newcommand{\cLadGSwrDelta}{+0.43}
\newcommand{\cLadGSwrDeltaHi}{+0.68}
\newcommand{\cLadGSwrDeltaLo}{+0.12}

\newcommand{\cLadGSwrIqm}{0.67}

\newcommand{\cLadGSwrSteps}{87.5k}

\newcommand{\cLadHEpsFail}{1}

\newcommand{\cLadHPaSwrDelta}{$-$0.32}
\newcommand{\cLadHPaSwrDeltaHi}{$-$0.09}
\newcommand{\cLadHPaSwrDeltaLo}{$-$0.58}

\newcommand{\cLadHSwrFail}{2}

\newcommand{\cLadHSwrN}{5}

\providecommand{\cEpLenGoalMin}{\tbd{--}}
\providecommand{\cEpLenGoalMax}{\tbd{--}}
\providecommand{\cDeadFour}{\tbd{--}}
\providecommand{\cDeadFourUni}{\tbd{--}}
\providecommand{\cDeadSix}{\tbd{--}}
\providecommand{\cDeadSixUni}{\tbd{--}}
\providecommand{\cDeadEight}{\tbd{--}}
\providecommand{\cDeadEightUni}{\tbd{--}}

\icmltitlerunning{Latent Imagination for Parameterized Action Spaces}

\begin{document}
\twocolumn[
\icmltitle{Latent Imagination for Parameterized Action Spaces}
\icmlsetsymbol{equal}{*}
\begin{icmlauthorlist}
\icmlauthor{Atahan Cilan}{aff1}
\icmlauthor{\.{I}hsan Mert Muhac{\i}ro\u{g}lu}{aff1}
\icmlauthor{\"{U}mit Can Bekar}{aff1}
\end{icmlauthorlist}
\icmlaffiliation{aff1}{Turkish Aerospace} 
\icmlcorrespondingauthor{Atahan Cilan}{atahan.cilan@tai.com.tr}
\icmlkeywords{reinforcement learning, world models, parameterized actions}
\vskip 0.3in
]
\printAffiliationsAndNotice{}

\begin{abstract}
Parameterized-action MDPs (PAMDPs) ask an agent to choose a discrete action
together with continuous parameters whose meaning depends on that choice.
Model-free PAMDP methods act fast but need many interactions; model-based
methods are sample-efficient but plan at every decision, which makes them slow
to act. We introduce \ours{} (Hybrid Latent Imagination), a model-based agent
that learns a hybrid policy entirely in the latent imagination of a world model
and acts with a single forward pass. To address learning problems that the
PAMDP structure creates, \ours{} adds \cme{}, which gives every discrete action
a learning signal at every imagined state, and \swr{}, which keeps rare
successes in training. On seven standard PAMDP benchmarks, \ours{} is within
one published standard deviation of, or above, the published score of the strongest
model-free baseline in our comparison on all \cNmatch{} tasks, reaches 90\% of
that score with
\cEffHyarNoPre$\times$ fewer environment steps than that baseline
(\cEffHyar$\times$ when its reward-free pre-training is counted), and decides \cDlpaRatioCpu$\times$ faster than the model-based planner \dlpa{} at its published planner
settings. It is the only method in our comparison that is both more
sample-efficient than that baseline and fast enough for real-time control.
\end{abstract}

\section{Introduction}
\label{sec:intro}

\begin{figure}[t]
\centering
\includegraphics[width=\linewidth]{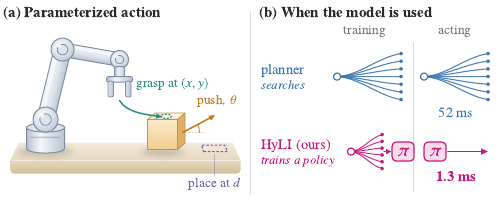}
\caption{\textbf{(a)}~A parameterized action: the robot chooses a primitive
(mode) and its arguments (parameter), whose meaning differs from primitive to
primitive. \textbf{(b)}~Both kinds of agent roll out a learned model. A planner
uses the rollouts to search for an action at every decision, during training
and when acting. \ours{} uses them only during training, to train a policy
$\pi$, and acts with one forward pass of $\pi$. Times: median per decision on
one CPU thread, \Goal{} benchmark (all tasks in \cref{tab:e6}).}
\label{fig:teaser}
\end{figure}

Many real-world control problems combine a discrete choice with continuous
parameters. A robot selects a manipulation primitive and its
arguments~\citep{dalal2021raps,nasiriany2022maple} (\cref{fig:teaser}a) or
switches between control
modes~\citep{neunert2020hybrid}; an automated vehicle decides whether to change
lanes and how~\citep{lin2024lanechange}; a UAV chooses between moving and
turning and sets the acceleration or the turn angle~\citep{feng2023uav}; a game
agent selects a skill and where to aim it~\citep{xiong2018pdqn}.
Parameterized-action MDPs (PAMDPs) formalize this structure~\citep{masson2016pamdp}:
at each step the agent chooses a mode $\modek\in\{1,\dots,\Kmodes\}$ and a
parameter $\paramz_\modek\in\mathbb{R}^{\paramdim{\modek}}$ whose dimension and
meaning may differ from mode to mode. Flattening the hybrid space into a
purely discrete or a purely continuous one---discretizing the parameters or
relaxing the mode---either loses precision or lets the agent mix incompatible
actions. PAMDP algorithms are
developed and compared on a common suite of simulated benchmarks, from robot
soccer and a platform game to multi-actuator navigation
tasks~\citep{masson2016pamdp,hausknecht2016pamdp,li2022hyar}.

Model-free PAMDP methods~\citep{xiong2018pdqn,fan2019hppo,li2022hyar} act with
a single forward pass of their networks but need many interactions. The model-based methods,
\dlpa{}~\citep{dlpa} and FLEXplore~\citep{flexplore}, learn a dynamics model
and report large gains in sample efficiency, but they select every action by
planning (\cref{fig:teaser}b): thousands of candidate sequences are rolled
through the model per decision, and the cost grows with the planning horizon
and, in the released implementation, with the number of modes. Diffusion policies such
as \chdp{}~\citep{chdp} run denoising chains per decision. Many of the
applications above, however, run in a control loop at tens to hundreds of
decisions per second.

We ask whether a model-based PAMDP agent can keep the sample efficiency of a
learned model without searching at decision time. World-model agents in the
Dreamer line~\citep{hafner2023dreamerv3} learn a policy inside imagined latent
trajectories and act with one forward pass. Carrying this to PAMDPs is not a
matter of swapping the action head, because a PAMDP is two coupled but
structurally different learning problems: the discrete choice has no unbiased
pathwise gradient, and the continuous choice can be differentiated through the model
only along the mode that was sampled, so a rarely chosen mode's parameters
are rarely trained and can starve.

We propose \ours{} (Hybrid Latent Imagination), which learns a hybrid
mode-and-parameter policy in imagination and gives each half of the action its
own gradient estimator. To keep rarely chosen modes learning, \cme{} (\CME{})
values every candidate mode with one imagined step; to keep rare successes in
training, \swr{} (\SWR{}) draws part of every batch from windows that contain
one. Our contributions are:
\begin{itemize}
  \item \ours{} (\cref{sec:method}), to our knowledge the first agent to
    bring latent imagination to the standard PAMDP benchmarks with a separate
    gradient estimator for each half of the action and \cme{}, compared there
    against planners. It acts with one forward pass of its policy, with one
    configuration whose per-task switches follow fixed rules, on tasks
    with 2 to 256 nominal modes. It
    is within one published standard deviation of, or above, the published
    score of the strongest model-free baseline in our comparison on all
    \cNmatch{} tasks, reaches 90\% of it with
    \cEffHyarNoPre$\times$ fewer interactions (geometric mean;
    \cEffHyar$\times$ with its reward-free pre-training counted) and
    decides \cDlpaRatioCpu$\times$ faster than \dlpa{} at its published
    settings (\cref{sec:exp:e1}): a balance of sample efficiency, final performance
    and decision speed that neither the model-free methods nor the planner
    reach.
  \item \Cme{} (\cref{sec:method:cme}), which trains the mode head with an
    expected policy gradient and gives every parameter head a gradient whether
    or not its mode was sampled. Along the cumulative ablation, it and \SWR{}
    account for most of the gain on the two sparse-success tasks
    (\cref{sec:exp:ablation}).
  \item A detailed analysis of actor designs for PAMDPs: a cumulative and a
    leave-one-out ablation, the direct adaptation of DreamerV3 with a relaxed
    continuous actor, and a comparison of shared and per-mode parameter heads
    that measures cross-mode gradient interference and its effect on
    performance (\cref{sec:exp:ablation,sec:exp:e2}). All experiments follow a
    reproducible protocol: every baseline runs from its official code (agents
    unchanged, drivers ported), every environment interaction is counted, and
    all methods are scored on the same window of interactions (\dlpa{} at its
    shorter converged budget; \cref{sec:exp:protocol},
    \cref{app:protocol,app:baselines}). Code will be released.
\end{itemize}

\section{Background}
\label{sec:background}

\paragraph{PAMDPs.}
\label{sec:bg:pamdp}
The action space of a PAMDP~\citep{masson2016pamdp} is
$\mathcal{A}=\bigcup_{\modek=1}^{\Kmodes}\{(\modek,\paramz_\modek)
: \paramz_\modek\in\mathcal{Z}_\modek\subseteq\mathbb{R}^{\paramdim{\modek}}\}$.
The dimension $\paramdim{\modek}$ may differ across modes (in \Goal{}, a kick
takes a 2-D target and each shot a 1-D target position on the goal line), and so may the \emph{semantics}
of $\paramz_\modek$: in \Goal{} the parameters of different modes are different
physical quantities (\emph{heterogeneous}), whereas in \HardMoveN{} all $2^n$
modes take the same scalar thrust (\emph{homogeneous}).

\paragraph{Latent imagination.}
\label{sec:bg:dreamer}
DreamerV3~\citep{hafner2023dreamerv3}, the latest agent of the Dreamer
line~\citep{hafner2019planet,hafner2020dreamer,hafner2021dreamerv2}, learns a
recurrent state-space model (RSSM) with deterministic state $\deter_t$ and categorical stochastic state
$\stoch_t$ from replayed sequences. An actor and a critic are trained purely on
trajectories imagined by the model; the critic regresses $\lambda$-returns and
the actor maximizes normalized returns plus an entropy bonus. At deployment only
the encoder, the recurrent core and the actor run. We change the action
interface, the actor and its gradient estimators; the world model and critic
follow DreamerV3 with reduced sizes, re-implemented in PyTorch
(\cref{app:repro,app:baselines:ours}).

\paragraph{Planning in PAMDPs.}
\label{sec:bg:planning}
\dlpa{}~\citep{dlpa} learns hybrid-action-conditioned dynamics, reward and
termination models and selects each action by MPPI~\citep{williams2017mppi}:
for $I$ iterations it samples $N$ sequences of length $L$, each step a mode and
a per-mode Gaussian parameter, rolls them through the model and refits the
sampling distributions. With the published settings ($I{=}6$, $N{=}1000$,
$L\in\{5,8,10\}$) one decision costs $I\cdot N\cdot L =$
30\,000--60\,000 model steps. This count does not depend on $\Kmodes$; the
released implementation, however, loops over the modes serially
(\cref{app:baselines}), so its decision time also grows with $\Kmodes$.

\section{Method}
\label{sec:method}

\Cref{fig:method} gives an overview of \ours{}, from acting to training. The world model is a DreamerV3 RSSM
whose action input encodes the hybrid action; the actor has a discrete mode head
and per-mode parameter heads (\cref{sec:method:actor}), trained in imagination
with a different gradient estimator for each half (\cref{sec:method:grad}) and
with \cme{} (\cref{sec:method:cme}). Nothing in
\cref{sec:method:grad,sec:method:cme,sec:method:train} runs at deployment.

\begin{widefigure}
  \centering
  \includegraphics[width=\linewidth]{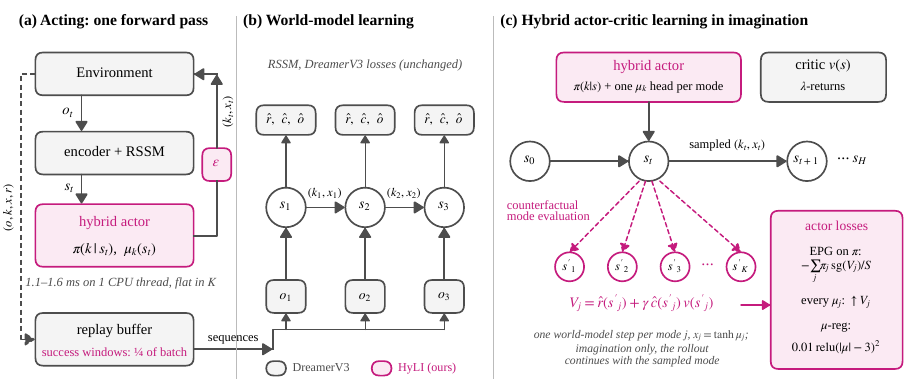}
  \caption{Overview. \textbf{(a)}~Acting is one forward pass through the encoder,
  the recurrent core and the hybrid actor; on \Goal{} and \HardGoal{} the mode is
  occasionally replaced by a uniform one early in training ($\varepsilon$).
  \textbf{(b)}~The world model is a DreamerV3 RSSM trained on replayed sequences,
  a quarter of each batch drawn from windows that contain a success (\SWR{}).
  \textbf{(c)}~The actor and critic learn on imagined rollouts that continue with
  the sampled mode. From every imagined state, \CME{} (dashed) takes one model
  step per candidate mode $j$ and values it, $V_j$; the values train the mode
  head by the expected policy gradient and every parameter head by its own
  $V_j$. Magenta: parts that \ours{} adds for PAMDPs; grey: unchanged DreamerV3
  components.}
  \label{fig:method}
\end{widefigure}

\subsection{Hybrid actor}
\label{sec:method:actor}
\label{sec:method:repr}
The world model receives
$[\,\mathrm{onehot}(\modek)\;|\;\paramz_1\;|\;\dots\;|\;\paramz_\Kmodes\,]$
with every slice but the selected mode's set to zero, so no two modes alias an
input unit; on \HardMoveN{}, where every mode has the same single parameter,
the smaller $[\,\mathrm{onehot}(\modek)\;|\;\paramz_\modek\,]$ is equivalent
(\cref{app:protocol:repr}). The actor maps the latent feature
$s=(\deter,\stoch)$ to a categorical $\pi(\modek\mid s)$ (1\% uniform mixing)
and, per mode, a Gaussian over pre-squash parameters with mean $\mu_\modek(s)$;
the executed parameter is $\tanh$ of a sample. The parameter head is either
\texttt{sub}, one MLP per mode (no cross-mode interference by construction), or
\texttt{multi}, a shared trunk with one output row per mode.

Whether the parameter heads should share weights is not obvious. When the
parameters of different modes are different physical quantities, an update that
improves one mode's parameter (a kick target) can disturb another's (a shot
position) through a shared trunk, a form of gradient interference; when they are
the same quantity, sharing may instead help. To measure how this choice
affects performance, we implement both heads and compare them in
\cref{sec:exp:e2}. By default we use \texttt{multi} where every mode has the
same single parameter (\HardMoveN{}) and \texttt{sub} on the other tasks.

\subsection{Split gradient estimators}
\label{sec:method:grad}
The continuous half is differentiable through the model: the parameter loss is
the negative normalized $\lambda$-return, backpropagated through the imagined
dynamics as in DreamerV3's continuous control; in this term the critic's
bootstrap values are detached, so the gradient flows through the predicted
rewards and continuations (\cref{app:repro}). The discrete half has no
unbiased pathwise gradient (straight-through and Gumbel-softmax
estimators~\citep{bengio2013straight,jang2017gumbel} are biased). Without
\CME{} it is trained with the score-function estimator
$-\log\pi(\modek_t\mid s_t)\,\mathrm{sg}(A_t)$, with stop-gradient
$\mathrm{sg}$ and advantage $A_t$, the normalized $\lambda$-return minus the
critic's value; in the full method \CME{}
replaces this estimator by an expected policy gradient
(\cref{sec:method:cme}), so the score-function estimator is used only in the
first rows of the ablation ladder (\cref{sec:exp:ablation}). In both cases the
one-hot of the sampled mode enters the world model \emph{detached}, so the
return gradient does not reach the mode logits a second time through a relaxed
one-hot. Both halves carry
an entropy bonus of $3\cdot10^{-4}$ on every task; for the parameters it is the
entropy of the selected mode's pre-squash Gaussian.

\subsection{Counterfactual mode evaluation}
\label{sec:method:cme}
The score-function estimator learns from the one mode that was sampled, and the
pathwise gradient improves that mode's parameters only. A rarely sampled mode is
rarely improved, can look worse and be sampled even less. \CME{} is
designed to break this loop with the world model we already have. At every imagined state $s_t$ and for each
of the $\Ncand$ candidate modes $j$ (the set is defined below) we take one imagined step with that mode and its
deterministic parameter,
\begin{align}
  s'_j &= f\big(s_t,\,(j,\tanh\mu_j(s_t))\big), \nonumber\\
  V_j  &= \hat r(s'_j) + \discount\,\hat c(s'_j)\,v(s'_j),
  \label{eq:cme}
\end{align}
where $f$ is one step of the RSSM prior (one sample) and $\hat r$, $\hat c$
and $v$ are the existing reward, continuation and critic heads. Because the
parameter is the mean, $V_j$ is a biased estimate of the mode's value under the
stochastic policy. The imagined
trajectory still continues with the sampled mode, so the extra cost is
$\imhorizon\cdot\Ncand$ one-step evaluations per start state, not a tree. The
values are used twice: the mode head receives, in place of the
score-function estimator, the expected policy
gradient~\citep{ciosek2018epg} $-\sum_j \pi_j\,\mathrm{sg}(V_j)/S$, which has
zero variance in the mode choice ($S$: DreamerV3's return scale,
\cref{app:protocol:train});
every candidate's parameter head ascends its own value,
$-\tfrac1{\Ncand}\sum_j V_j/S$, whether or not the mode was sampled. This
term differentiates through $v$ but does not update the critic. With
\texttt{sub} heads, the pathwise term reaches mode $j$'s parameters only when
$j$ is sampled, with weight $\pi_j(s_t)$ in expectation; this term gives every
candidate weight $1/\Ncand$, so the heads stay isolated and every candidate is trained at every imagined
state. The sampled mode's parameters receive this gradient
and the pathwise one with equal weight. The candidate set is all modes when
$\Kmodes\le 16$; otherwise the 8 most probable plus 8 drawn uniformly without
replacement from the rest, with $\pi$ renormalized over the set, a biased
approximation whose cost does not grow with $\Kmodes$.

\subsection{Replay, exploration and regularization}
\label{sec:method:train}
Three single-switch components target failures we observed on \Goal{} and
\HardGoal{}, where a random policy almost never scores; \cref{sec:exp:ablation}
adds and removes them one at a time. \Swr{} (\SWR{}) draws a quarter of every
batch from windows that contain a success event (on the MPE tasks, a step with
reward above 4, the success rule of the benchmark), so that rare early
successes are not diluted; it is a simple form of prioritized
replay~\citep{schaul2016per}, close in spirit to self-imitation
learning~\citep{oh2018sil}. Intuitively, \SWR{} keeps successful experience in view: the
world model, and the imagined rollouts the actor learns from, continue to
represent success while it is still rare. \Epsmode{} replaces the policy's mode
by a uniform one with a probability that decays from 0.1 to 0 over the first
$5\cdot10^4$ steps after the prefill; only the replay data change. Finally, a $\tanh$-squashed
parameter whose mean saturates receives almost no gradient and can stay at its
bound, a failure that \citet{hausknecht2016pamdp} documented for PAMDPs and
prevented by inverting gradients near the bounds. We use a simpler penalty:
\mureg{} adds $0.01\cdot\mathrm{relu}(|\mu_\modek|-3)^2$ to the actor loss,
on the bounded mean (\cref{tab:hyper}), which is zero for $|\mu|\le3$ and pulls a saturating mean back. Replay,
imagination and prefill details are in \cref{app:protocol:train}.

\section{Experiments}
\label{sec:exp}
We first compare \ours{} with all baselines on final score, decision time and
sample efficiency (\cref{sec:exp:e1}), then study its components
(\cref{sec:exp:ablation}) and the design of its parameter head
(\cref{sec:exp:e2}).

\subsection{Protocol}
\label{sec:exp:protocol}
\paragraph{Tasks, arms and scores.} We use the PAMDP suite of
\citet{li2022hyar}: \Platform{}, \Goal{}, \HardGoal{}, \CatchPoint{} and
\HardMoveN{}, $n\in\{4,6,8\}$, the last two built on the multi-agent particle
environment (MPE)~\citep{lowe2017mpe}, with $\Kmodes$ from 2 modes on
\CatchPoint{} to 256 on \HardMove{8} (\cref{tab:tasks}). We call each compared
method an \emph{arm} and each arm--task pair a \emph{cell}. The baselines
\pdqn{}~\citep{xiong2018pdqn}, \patdd{}~\citep{hausknecht2016pamdp,li2022hyar},
\hhqn{}~\citep{fu2019hhqn}, \hppo{}~\citep{fan2019hppo} and
\hyar{}~\citep{li2022hyar}, all but \hppo{} in their TD3
variants~\citep{fujimoto2018td3}, run the reference agents unchanged with
ported drivers and their per-task settings; the released \hhqn{}-TD3 differs
from TD3 in ways we keep (\cref{app:env:hhqn}). \dlpa{} is ported from its
official code and verified bit-exact against it; \pdqn{} keeps its hand-coded parameter prior on
\Platform{} and \Goal{}. Every baseline runs with the per-task settings its authors published for
these tasks, without re-tuning; while adding
our components we tried variants on every trained task except \HardMove{8},
mostly on \Goal{} and \HardGoal{}, then fixed one configuration for every
experiment; \HardMove{8} was not used to choose any component or setting. Every arm is trained with the same
five seeds (0--4) on every task. A run's score is the
mean of its final five evaluations (100 greedy episodes each, in a separate
environment); the metric is the reference one, return on \Platform{} and
success rate elsewhere. Every environment interaction counts, \hyar{}'s
reward-free pre-training included, and all arms are scored on the same window:
their first $B$ interactions, $B$ our budget on the task (\dlpa{} at its
shorter converged budget). \dlpa{} plans at
every training step, so a run at the reference budget would take one to
several weeks per seed. It runs at the budget of its own paper; where it is not
within one published standard deviation of its published return there, we extend the run
until it converges. Details and every deviation from the
reference code: \cref{app:protocol,app:baselines}.

\paragraph{Metrics.} Sample efficiency is the number of environment steps a
run needs to first reach 90\% of \hyar{}'s published score (median over seeds;
a seed that never reaches it is censored at its budget). We compare decision
rates against 50\,Hz, a common rate for learned low-level control on
physical systems: legged robots run their learned policies at
50\,Hz~\citep{rudin2022walk} and 200\,Hz~\citep{hwangbo2019agile}, and
autopilots typically sample at 100\,Hz~\citep{bohn2019fixedwing}. A racing agent that competes against humans in
real time in a driving simulator acts at 10\,Hz~\citep{wurman2022sophy}, so we
also state the result at 10\,Hz.

\subsection{Results}
\label{sec:exp:e1}
\label{sec:exp:e6}
\begin{widetable}
\centering
\small
\setlength{\tabcolsep}{3pt}
\begin{tabular}{lcccccccc}
\toprule
Task & Budget & \ours{} (ours) & \dlpa$^\dagger$ & \hyar & \pdqn$^\ddagger$ & \hhqn & \patdd & \hppo \\
\midrule
\Platform & 2e5 & \textbf{1.00\,$\pm$\,0.00} & 0.97\,$\pm$\,0.03 & 0.98\,$\pm$\,0.01 & 0.84\,$\pm$\,0.13 & 0.42\,$\pm$\,0.22 & 0.92\,$\pm$\,0.03 & 0.81\,$\pm$\,0.03 \\
\Goal & 3e5 & \textbf{0.77\,$\pm$\,0.03} & 0.58\,$\pm$\,0.29 & 0.74\,$\pm$\,0.02 & 0.61\,$\pm$\,0.20 & 0.00\,$\pm$\,0.00 & 0.00\,$\pm$\,0.00 & 0.00\,$\pm$\,0.00 \\
\HardGoal & 3e5 & \textbf{0.79\,$\pm$\,0.04} & 0.34\,$\pm$\,0.28 & 0.59\,$\pm$\,0.08 & 0.07\,$\pm$\,0.08 & 0.02\,$\pm$\,0.04 & 0.14\,$\pm$\,0.18 & 0.00\,$\pm$\,0.00 \\
\CatchPoint & 5e5 & \textbf{0.94\,$\pm$\,0.00} & 0.92\,$\pm$\,0.03 & 0.70\,$\pm$\,0.18 & 0.92\,$\pm$\,0.05 & 0.33\,$\pm$\,0.17 & 0.59\,$\pm$\,0.17 & 0.72\,$\pm$\,0.06 \\
\HardMove{4} & 5e5 & 0.91\,$\pm$\,0.05 & 0.94\,$\pm$\,0.01 & \textbf{0.95\,$\pm$\,0.01} & 0.30\,$\pm$\,0.05 & 0.55\,$\pm$\,0.19 & 0.49\,$\pm$\,0.18 & 0.64\,$\pm$\,0.12 \\
\HardMove{6} & 5e5 & \textbf{0.95\,$\pm$\,0.03} & \textit{0.94\,$\pm$\,0.01} & \textit{0.93\,$\pm$\,0.04} & 0.34\,$\pm$\,0.06 & 0.17\,$\pm$\,0.05 & 0.07\,$\pm$\,0.04 & 0.59\,$\pm$\,0.19 \\
\HardMove{8} & 5e5 & 0.91\,$\pm$\,0.06 & \textbf{0.93\,$\pm$\,0.01} & 0.89\,$\pm$\,0.06 & 0.14\,$\pm$\,0.05 & 0.05\,$\pm$\,0.04 & 0.08\,$\pm$\,0.03 & 0.14\,$\pm$\,0.05 \\
\bottomrule
\end{tabular}

\caption{Final score on a common window: every arm is scored on its first $B$
environment interactions ($B$: Budget column), mean\,$\pm$\,std over 5 seeds of
the mean of the final five evaluations (return on \Platform{}, success rate
elsewhere). $^\dagger$At its converged, shorter budget. $^\ddagger$With a
hand-coded parameter prior. Bold: best mean; italic: mean at least the best mean minus the best arm's
standard deviation, both computed before rounding. Std: population standard
deviation over seeds, as in \citet{li2022hyar}. Full budgets and learning
curves: \cref{app:extra}.}
\label{tab:e1}
\end{widetable}

\begin{table}[t]
\centering
\small
\setlength{\tabcolsep}{3pt}
\begin{tabular}{lrccccr}
\toprule
Task & $\Kmodes$ & \oursshort & \dlpa & \hyar & \pdqn & ratio \\
\midrule
\Platform & 3 & 1.29 & 108 & 0.343 & 0.11 & 84$\times$ \\
\Goal & 3 & 1.3 & 51.6 & 0.347 & 0.112 & 40$\times$ \\
H.\,Goal & 11 & 1.56 & 62 & 0.347 & 0.107 & 40$\times$ \\
Catch & 2 & 1.28 & 52.4 & 0.337 & 0.106 & 41$\times$ \\
\HM{4} & 16 & 1.09 & 46.2 & 0.345 & 0.106 & 42$\times$ \\
\HM{6} & 64 & 1.1 & 95.9 & 0.346 & 0.109 & 87$\times$ \\
\HM{8} & 256 & 1.13 & 323 & 0.357 & 0.124 & 286$\times$ \\
\HM{10} & 1024 & 1.29 & 1145 & 0.381 & 0.222 & 888$\times$ \\
\bottomrule
\end{tabular}

\caption{Milliseconds per decision on one CPU thread, batch size 1, median of
1\,000 decisions: for \ours{} the encoder, recurrent step and actor; for
\dlpa{} a full MPPI plan. Same machine for all arms; GPU and all baselines in
\cref{app:extra}. Ratio: \dlpa{} over ours.}
\label{tab:e6}
\end{table}

\begin{figure}[t]
\centering
\includegraphics[width=\linewidth]{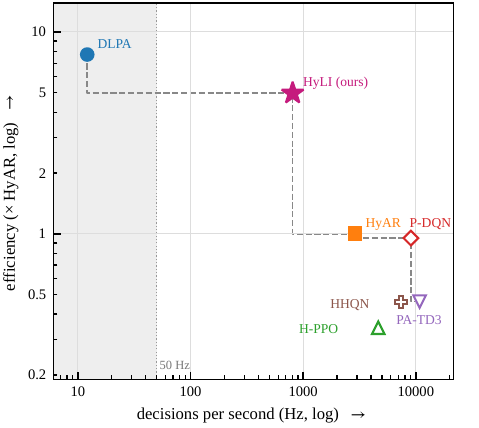}
\caption{Sample efficiency against decision rate, both higher-is-better.
$x$: decisions per second on one CPU thread (\cref{tab:e6} and
\cref{tab:e6-full}). $y$: \hyar{}'s steps to the threshold, pre-training
included, divided by the arm's (\cref{tab:efficiency}), geometric mean over
tasks; open markers are upper bounds (a task median is censored).
Shaded: below 50\,Hz. Dashed: Pareto front. Per task: \cref{app:extra}.}
\label{fig:pareto}
\end{figure}

\begin{widetable}
\centering
\small
\setlength{\tabcolsep}{3pt}
\begin{tabular}{lcccccccc}
\toprule
Task & \ours{} (ours) & \dlpa$^\dagger$ & \hyar & \hyar$^\S$ & \pdqn$^\ddagger$ & \hhqn & \patdd & \hppo \\
\midrule
\Platform{} (0.88) & 4.5k & \textbf{1.5k} & 25k & 14k & 9k & $>$200k\,\scriptsize(2/5) & 21.5k & $>$200k\,\scriptsize(0/5) \\
\Goal{} (0.70) & \textbf{45k} & 47k\,\scriptsize(4/5) & 85.8k & 65k & 31.5k\,\scriptsize(4/5) & $>$300k\,\scriptsize(0/5) & $>$300k\,\scriptsize(0/5) & $>$300k\,\scriptsize(0/5) \\
\HardGoal{} (0.54) & \textbf{29.5k} & 29k\,\scriptsize(3/5) & 105k & 84.5k & $>$300k\,\scriptsize(0/5) & $>$300k\,\scriptsize(0/5) & $>$300k\,\scriptsize(2/5) & $>$300k\,\scriptsize(0/5) \\
\CatchPoint{} (0.81) & 20k & \textbf{10k} & 221k & 115k & 105k & $>$500k\,\scriptsize(0/5) & $>$500k\,\scriptsize(2/5) & 432k\,\scriptsize(4/5) \\
\HardMove{4}{} (0.84) & 27.5k & \textbf{15k} & 185k & 40k & $>$500k\,\scriptsize(2/5) & 55k & $>$500k\,\scriptsize(1/5) & $>$500k\,\scriptsize(2/5) \\
\HardMove{6}{} (0.83) & 35k & \textbf{22.5k} & 212k & 67.5k & 340k\,\scriptsize(4/5) & $>$500k\,\scriptsize(0/5) & $>$500k\,\scriptsize(0/5) & $>$500k\,\scriptsize(0/5) \\
\HardMove{8}{} (0.80) & 62.5k & \textbf{47.5k} & 275k & 130k & $>$500k\,\scriptsize(0/5) & $>$500k\,\scriptsize(0/5) & $>$500k\,\scriptsize(0/5) & $>$500k\,\scriptsize(0/5) \\
\bottomrule
\end{tabular}

\caption{Environment steps to reach 90\% of \hyar{}'s published score
(threshold in brackets), median over 5 seeds; (reached/5) when not every seed
reaches it; $>$: censored at the arm's budget. Bold: fewest steps among the
arms whose five seeds all reach the threshold, so that a fast median over
partly failed seeds is not ranked first. \hyar{}'s steps include its
reward-free pre-training; $^\S$: \hyar{} without it (not ranked).}
\label{tab:efficiency}
\end{widetable}

\begin{widefigure}
\centering
\includegraphics[width=\linewidth]{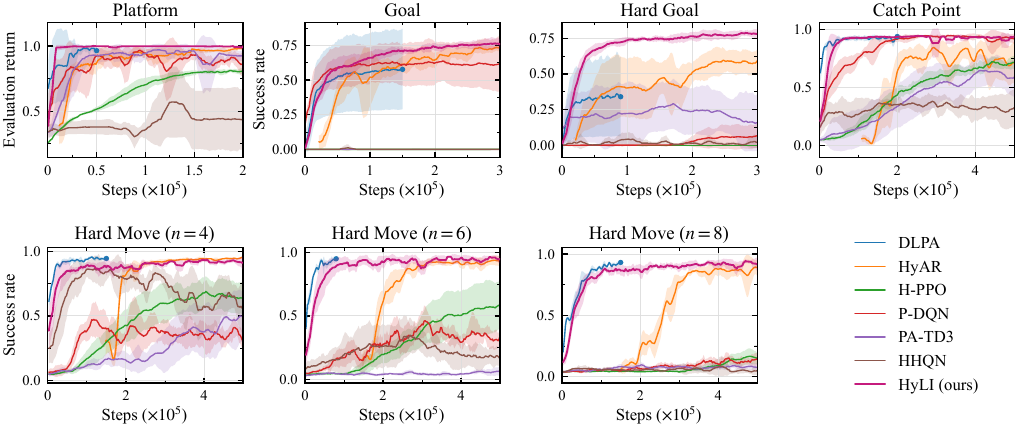}
\caption{Learning curves in the metric of \cref{tab:e1}: evaluation success
rate (return on \Platform{}), mean\,$\pm$\,std over 5 seeds. Each panel ends
at our budget; arms with longer budgets are drawn up to it, and a dot marks the
end of a shorter one (\dlpa{}). Every interaction is on the step axis,
\hyar{}'s reward-free pre-training included; its curve therefore starts later.
Episode return: \cref{fig:curves-return}.}
\label{fig:curves}
\end{widefigure}

\paragraph{Final score.} \ours{} has the highest mean on \cNbestMean{} of the
7 tasks (\cref{tab:e1}) and no failing seed. The exceptions are
\HardMove{4}, where \hyar{} is ahead within the window, and \HardMove{8}, where
\dlpa{} is. Against the published results of \citet{li2022hyar} it is within
one published standard deviation or above on all \cNmatch{} tasks and more
than one above on \cNabove{} (\Platform{}, \HardGoal{} and \CatchPoint{}).
\chdp{} reports higher success on \Goal{}, \HardGoal{} and \HardMove{4} under its own
protocol; without released code we do not rank against it (\cref{app:published}).
\Goal{} and \HardGoal{} separate the arms: \hhqn{} and \hppo{} stay near zero
success on both and \patdd{} on \Goal{}, at the deceptive optimum of shooting
from the start position, and some \dlpa{} seeds lock into the same
optimum and never leave it.

\paragraph{Decision time.} One decision of \ours{} takes \cOursMsCpu\,ms on one
CPU thread (at least \cOursMinHzCpu\,Hz on every task, \cref{tab:e6}). \dlpa{}
is \cDlpaRatioCpu$\times$ slower (geometric mean over the seven trained
tasks) and reaches at most \cDlpaMaxHzCpu\,Hz, \cDlpaMaxHzGpu\,Hz on a GPU, and even a
10\,Hz loop excludes it on \Platform{}, \HardMove{8} and \HardMove{10}. The
cost of every policy stays nearly flat from $\Kmodes{=}2$ to
$\Kmodes{=}1024$, whereas the released planner's grows with its horizon and,
because it loops over modes serially (\cref{app:baselines}), with $\Kmodes$,
from \cDlpaRatioMinCpu$\times$ to
\cDlpaRatioMaxCpu$\times$ ours on the trained tasks and to
\cDlpaHmTenMsCpu\,ms on \HardMove{10}. The model-free policies are
\cMfRatioMinCpu{} to \cMfRatioMaxCpu$\times$ faster than ours. The extra
computation of \ours{} is paid during training instead: \CME{} multiplies the
update time by 1.4 to 2.6 (\cref{app:cost}).

\paragraph{Sample efficiency.} \Cref{fig:pareto} summarizes the trade-off. The
model-free arms decide fastest, but none is more sample-efficient than \hyar{}
in geometric mean, and several never reach the threshold. Counting only arms whose five seeds all reach the
threshold (\cref{tab:efficiency}), \dlpa{} needs the fewest interactions on
\cNfastestDlpa{} of the 7 tasks; at its published planner settings it decides
below 50\,Hz on every task.
\ours{} lies on the front between them and is the only arm both above 50\,Hz
and more sample-efficient than \hyar{}: \cEffHyarNoPre$\times$ fewer steps
than \hyar{} without its reward-free pre-training (geometric mean;
\cEffHyarNoPreMin$\times$ to \cEffHyarNoPreMax$\times$ per task) and
\cEffHyar$\times$ with it. By the same count it is the most sample-efficient
arm on \Goal{} and \HardGoal{}, where some \dlpa{} seeds never reach the
threshold, and the most sample-efficient of the arms that act without a
planner on all \cNfastestOursRealtime{} tasks.

\subsection{Ablations}
\label{sec:exp:ablation}
We study the components on \Goal{} and \HardGoal{}, the two sparse-success
tasks on which most of the development took place (five seeds per cell). The cumulative
ladder (\cref{fig:ladder-main,tab:ladder}) adds them one at a time to the bare
hybrid actor, in a fixed order; two final rows replace the
hybrid actor with a relaxed continuous one, which yields the direct, naive
adaptation of DreamerV3. The leave-one-out ablation (\cref{tab:ablation})
removes one component from the full method.

\begin{widefigure}[t]
\centering
\includegraphics[width=0.46\linewidth,trim=0 157 0 0,clip]{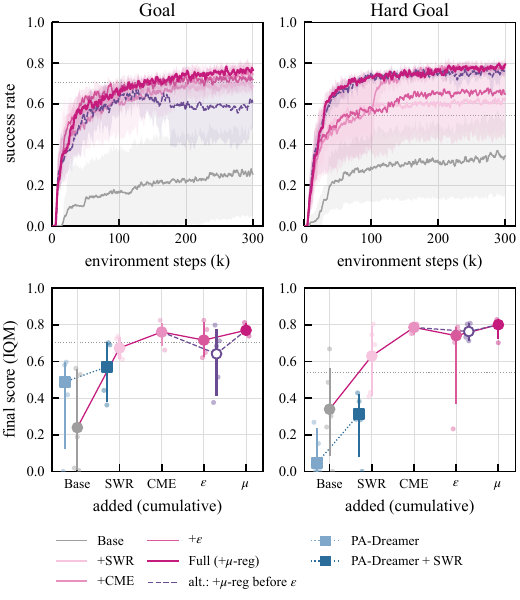}\hfill
\includegraphics[width=0.46\linewidth,trim=0 40 0 131,clip]{figures/ladder}\\[2pt]
\includegraphics[width=0.46\linewidth,trim=0 0 0 249,clip]{figures/ladder}
\caption{Cumulative ladder. Left: evaluation success rate, mean over 5 seeds
with a 95\% pointwise bootstrap band; dotted: 90\% of \hyar{}'s published
score. Right: inter-quartile mean of the final score with its 95\% bootstrap
interval and the individual seeds after each component is added; hollow
marker: the alternative path ($+$\muregshort{} before \epsshort{}); squares:
PA-Dreamer without and with \SWR{}. Each pair shows \Goal{} on the left and
\HardGoal{} on the right.}
\label{fig:ladder-main}
\vspace{6pt}
{\small
\setlength{\tabcolsep}{3pt}
\begin{tabular}{lcccccc}
\toprule
 & \multicolumn{3}{c}{\Goal} & \multicolumn{3}{c}{\HardGoal} \\
\cmidrule(lr){2-4} \cmidrule(lr){5-7}
 & score & $\Delta$ & steps & score & $\Delta$ & steps \\
\midrule
Base (all off) & 0.24 (0.01) [3] & -- & $>$300k\,\scriptsize(1/5) & 0.34 (0.00) [4] & -- & $>$300k\,\scriptsize(2/5) \\
\quad $+$\SWR{} & 0.67 (0.61) [0] & $+$0.43\,\scriptsize[$+$0.12, $+$0.68] & 87.5k & 0.63 (0.41) [2] & $+$0.29\,\scriptsize[$-$0.04, $+$0.58] & 34.5k \\
\quad $+$\CME{} & 0.76 (0.66) [0] & $+$0.09\,\scriptsize[$+$0.00, $+$0.16] & 63.5k & 0.79 (0.75) [0] & $+$0.16\,\scriptsize[$-$0.00, $+$0.37] & 30k \\
\quad $+$\epsshort{} & 0.72 (0.62) [0] & $-$0.04\,\scriptsize[$-$0.14, $+$0.06] & 40k & 0.74 (0.23) [1] & $-$0.05\,\scriptsize[$-$0.41, $+$0.01] & 24k\,\scriptsize(4/5) \\
\quad $+$\muregshort{} (full method) & 0.77 (0.74) [0] & $+$0.05\,\scriptsize[$-$0.03, $+$0.14] & 45k & 0.80 (0.70) [0] & $+$0.06\,\scriptsize[$-$0.02, $+$0.42] & 29.5k \\
\midrule
\quad alt.: $+$\muregshort{} before \epsshort{} & 0.64 (0.38) [1] & $-$0.12\,\scriptsize[$-$0.34, $+$0.01] & 55.5k & 0.76 (0.71) [0] & $-$0.02\,\scriptsize[$-$0.07, $+$0.04] & 24k \\
\midrule
PA-Dreamer (relaxed actor) & 0.49 (0.00) [3] & $+$0.25\,\scriptsize[$-$0.24, $+$0.52] & $>$300k\,\scriptsize(1/5) & 0.05 (0.00) [5] & $-$0.29\,\scriptsize[$-$0.56, $+$0.01] & $>$300k\,\scriptsize(1/5) \\
\quad $+$\SWR{} & 0.57 (0.36) [2] & $-$0.10\,\scriptsize[$-$0.29, $+$0.05] & 161k\,\scriptsize(3/5) & 0.31 (0.00) [5] & $-$0.32\,\scriptsize[$-$0.58, $-$0.09] & 187.5k\,\scriptsize(3/5) \\
\bottomrule
\end{tabular}
}
\captionof{table}{Cumulative ladder, 5 seeds per row: inter-quartile mean of the final
score (worst seed) [seeds below 0.5]; $\Delta$: change against the row above
with 95\% bootstrap interval (alternative path: against $+$\CME{}; PA-Dreamer:
against the hybrid row with the same components, i.e.\ the effect of the
actor; computed before rounding); steps: median steps to 90\%
of \hyar{}'s published score. The last ladder row is the main-comparison
cell.}
\label{tab:ladder}
\end{widefigure}

\paragraph{Cumulative ladder.} On \Goal{} the bare agent fails on
\cLadGBaseFail{} of \cLadGBaseN{} seeds. \SWR{} turns it into a learning agent
(\cLadGSwrDelta{}, interval \cLadGSwrDeltaLo{} to \cLadGSwrDeltaHi{}); \CME{}
raises the score again (\cLadGCmeDelta{}, interval \cLadGCmeDeltaLo{} to
\cLadGCmeDeltaHi{}) and shortens the \emph{take-off}, the steps to the
threshold, from \cLadGSwrSteps{} to
\cLadGCmeSteps{} steps; \epsshort{} shortens it to \cLadGEpsSteps{} without
raising the final score. On \HardGoal{} \SWR{} makes every seed take off, but
\cLadHSwrFail{} of \cLadHSwrN{} then stall in a \emph{parameter-saturation
trap}: early in training a parameter mean saturates the $\tanh$, its gradient
vanishes, and the success rate stays flat for the rest of the run. Adding
\CME{} brings every seed above 0.5, including seeds that enter the trap and
later leave it.
\epsshort{} re-opens the trap on \cLadHEpsFail{} seed and \muregshort{} closes
it. Along the ladder, \SWR{} and \CME{} account for most of the gain, \epsshort{} buys a faster
take-off on \Goal{}, and \muregshort{} is the guard \epsshort{} needs.

\paragraph{Relaxed actor.} The direct adaptation of DreamerV3 keeps its
continuous actor and gives it the relaxed action of
\patdd{}~\citep{hausknecht2016pamdp}: one vector with a score per mode and
every mode's parameters, executed as the arg-max mode.
We train this PA-Dreamer with everything else unchanged, with and without
\SWR{}. On \HardGoal{} it fails on every seed either way, \cLadHPaSwrDelta{}
against the hybrid actor with \SWR{} (interval \cLadHPaSwrDeltaLo{} to
\cLadHPaSwrDeltaHi{}). On \Goal{} neither actor learns reliably without \SWR{}.
Adding \SWR{} raises the hybrid actor from \cLadGBaseIqm{} to \cLadGSwrIqm{}
(\cLadGSwrDelta{}) but PA-Dreamer only from \cLadGPaBaseIqm{} to
\cLadGPaSwrIqm{} (\cLadGPaSwrSwrDelta{}), so with \SWR{} PA-Dreamer ends below
the hybrid actor (\cLadGPaSwrDelta{}, \cref{tab:ladder}). PA-Dreamer fails by plateaus in the mode choice---shooting
from the start or dribbling without shooting---not by saturation.

\begin{table}[t]
\centering
\small
\setlength{\tabcolsep}{3pt}
\begin{tabular}{lcccc}
\toprule
 & \multicolumn{2}{c}{\Goal} & \multicolumn{2}{c}{\HardGoal} \\
\cmidrule(lr){2-3} \cmidrule(lr){4-5}
 & score & steps & score & steps \\
\midrule
Full method & 0.77 (0.74) [0] & 45k & 0.80 (0.70) [0] & 29.5k \\
\midrule
\quad $-$\SWR{} & 0.76 (0.69) [0] & 47.5k & 0.80 (0.75) [0] & 42.5k \\
\quad $-$\CME{} & 0.73 (0.63) [0] & 46k & 0.76 (0.62) [0] & 23k \\
\quad $-$\muregshort{} & 0.72 (0.62) [0] & 40k & 0.74 (0.23) [1] & 24k\,\scriptsize(4/5) \\
\quad $-$\epsshort{} & 0.64 (0.38) [1] & 55.5k & 0.76 (0.71) [0] & 24k \\
\bottomrule
\end{tabular}

\caption{Leave-one-out: inter-quartile mean of the final score (worst seed)
[seeds below 0.5], and median steps to 90\% of \hyar{}'s published score.
5 seeds per row. Curves: \cref{app:extra:ablation}.}
\label{tab:ablation}
\end{table}

\paragraph{Leave-one-out.} No single removal lowers the final score as much as
the ladder steps do (\cref{tab:ablation}), because the components overlap, as
the ladder shows. The effect of a removal shows in the worst seed and the
take-off: without
\epsshort{} one \Goal{} seed fails, without \muregshort{} one \HardGoal{} seed
stays in the saturation trap, without \SWR{} \HardGoal{} takes off later, and
without \CME{} the worst seed drops on both tasks.

\subsection{Parameter-head isolation}
\label{sec:exp:e2}
\begin{figure}[t]
\centering
\includegraphics[width=0.9\linewidth]{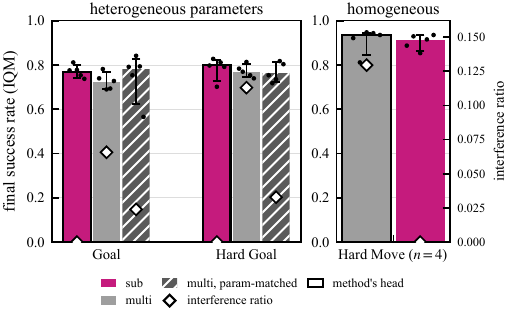}
\caption{Parameter heads. Bars: inter-quartile mean of the final score, 95\%
bootstrap interval, 5 seeds (dots); black outline: the head the method uses.
Diamonds (right axis): cross-mode interference ratio (zero for \texttt{sub};
\cref{app:protocol:interference}). Matched: \texttt{multi} widened to the
parameter count of \texttt{sub}. Scores: \cref{tab:e2}.}
\label{fig:e2}
\end{figure}
\Cref{fig:e2} compares the per-mode \texttt{sub} head with a shared
\texttt{multi} head and with a \texttt{multi} head widened to the parameter
count of \texttt{sub}. On \HardGoal{} the shared head reaches the score of
\texttt{sub}, although its measured cross-mode interference stays clearly
above zero. On \Goal{} it is slightly lower only when its capacity is not
matched, and the widened head closes the gap. On \HardMove{4} both heads reach
the same score. Interference between heterogeneous parameters is therefore
present but does not reduce performance on these tasks; we attribute none of
our results to isolation, and the default head is within noise of its
alternatives.

\section{Related work}
\label{sec:related}

\paragraph{Model-free PAMDP methods.} \citet{masson2016pamdp} alternate between
the discrete and the parameter policy; PA-DDPG~\citep{hausknecht2016pamdp}, built
on DDPG~\citep{lillicrap2016ddpg}, and its TD3 variant
\patdd{}~\citep{li2022hyar} relax the hybrid action into one continuous vector;
P-DQN and
MP-DQN~\citep{xiong2018pdqn,bester2019mpdqn} output all modes' parameters and
value each mode with a Q-function; \hhqn{}~\citep{fu2019hhqn} chooses the mode
and then its parameters with two hierarchical networks; H-PPO~\citep{fan2019hppo},
built on PPO~\citep{schulman2017ppo}, and Hybrid
SAC~\citep{delalleau2019hybridsac} use separate discrete and continuous actors; \hyar{}~\citep{li2022hyar} runs TD3 in a learned
latent action space and defines the benchmark protocol we use;
\daac{}~\citep{daac2026} treats distribution parameters as actions. All act with a
single forward pass of their networks; none uses a model.

\paragraph{Model-based PAMDP methods and decision-time computation.}
\dlpa{}~\citep{dlpa} and FLEXplore~\citep{flexplore} plan with MPPI in a learned
model at every decision. \dlpa{} also reports that a Dyna-like variant, which
learns a policy from model rollouts, falls behind planning on \Platform{} and
\Goal{}, so we compare against \dlpa{} with planning, the stronger of the two;
our ablations (\cref{sec:exp:ablation}) show which failures such a
policy meets in a PAMDP and which components remove them. \chdp{}~\citep{chdp}
runs a discrete and a continuous diffusion policy per decision, and
KGRL~\citep{kgrl2026} prunes actions with hand-written rules and refines
parameters by gradient ascent on its Q-function at decision time. Closest to our
approach, RAPS~\citep{dalal2021raps} trains a Dreamer agent over robot
manipulation primitives with a categorical primitive choice and one vector of
all primitives' arguments, both trained with the score-function estimator; it
is evaluated on manipulation suites, not on the PAMDP benchmarks, and is not
compared with planners. \ours{} differs in its separate estimator for each
half of the action and in \CME{}.

\paragraph{Evaluating every action.} The expected policy
gradient~\citep{ciosek2018epg}, mean actor-critic~\citep{asadi2017mac} and
many-actions policy gradients~\citep{nauman2023manyactions} integrate over
several actions per state, the last also with a model; COMA~\citep{foerster2018coma}
uses a counterfactual baseline over one agent's actions; MuZero and Sampled
MuZero~\citep{schrittwieser2020muzero,hubert2021sampledmuzero} evaluate
candidates in a learned model at decision time. \CME{} combines one
model step per candidate mode with these estimators at training time only, so
that it costs nothing when acting. TD-MPC2~\citep{hansen2024tdmpc2} plans in
continuous action spaces only.

\section{Limitations}
\label{sec:limitations}
Our ablations cover \Goal{} and \HardGoal{} only, the tasks on which most of
the development took place; of the trained tasks only \HardMove{8} was held
out from development. Among the
arms whose seeds all reach the threshold, \dlpa{} still needs the fewest
interactions on \cNfastestDlpa{} of the 7 tasks, so the
sample efficiency of learning a policy in imagination leaves room for
improvement; a seed of \ours{} also trains for tens of hours
(\cref{app:compute}). The benchmark's episodes are
short (a trained \ours{} takes \cEpLenGoalMin--\cEpLenGoalMax{} decisions per
episode on \Goal{} and \HardGoal{}),
which limits what the recurrent state can be credited for, and \CME{}
evaluates a capped candidate set, a biased approximation for large $\Kmodes$;
we train up to $\Kmodes{=}256$. No code is available for \chdp{} or
FLEXplore; we report \chdp{}'s published results (\cref{app:published}) and
run neither.

\section{Conclusion}
\label{sec:conclusion}
Model-based PAMDP methods learn from far fewer interactions than model-free
ones but plan at every decision, which limits how fast they can act. \ours{} learns a
hybrid policy in a world model's latent imagination and acts with one forward
pass. \CME{} gives every candidate mode a learning
signal at every imagined state, and \SWR{} keeps rare successes in training. On seven benchmarks, \ours{} keeps comparable or better final performance
while combining model-based sample efficiency with a policy's decision speed:
\cEffHyarNoPre$\times$ fewer interactions than \hyar{} and decisions
\cDlpaRatioMinCpu{} to \cDlpaRatioMaxCpu$\times$ faster than \dlpa{}. Longer horizons, where a critic can
look beyond a planner's, are the next test.

\section*{Impact Statement}
This paper presents work whose goal is to advance the field of machine
learning, specifically reinforcement learning for actions that combine a
discrete choice with continuous parameters. Its practical consequence is that
such policies can be learned with fewer interactions and executed fast enough
for real-time control loops, as in robots, vehicles and aircraft. Our results
are on simulated benchmarks; success rates on them say nothing about the
safety of a learned controller on a physical system, which requires
verification and runtime safeguards beyond the scope of this work. We see no
other societal consequence that must be highlighted here.

\bibliographystyle{icml2026}
\bibliography{refs}

\newpage
\appendix
\crefalias{section}{appendix}
\crefalias{subsection}{appendix}
\crefalias{subsubsection}{appendix}
\onecolumn
\raggedbottom   
\section{Protocol and method details}
\label{app:protocol}

\paragraph{Evaluation.} Every arm is evaluated with the reference cadence
(every 500 steps on \Platform{}, \Goal{}, \HardGoal{}; every 2\,500 on the MPE
tasks), 100 greedy episodes per evaluation, in a separate environment under a
preserved random stream, so that evaluation never perturbs training. A run's
score is the mean of its final five evaluations; cells report mean\,$\pm$\,std
over seeds 0--4. Because a single late evaluation dip can move this score,
\cref{app:extra} repeats every cell with the mean over the last
$\min(5\cdot10^4,\,25\%\text{ of the budget})$ steps. Curves plot the per-task
metric of the tables; \cref{fig:curves-return} repeats \cref{fig:curves} in
episode return, the quantity the reference papers plot. \HardMove{10} is not trained, to keep the compute budget (its 1024
nominal modes collapse to 211 distinct motions, \cref{app:env:degeneracy}); it
appears in the latency table and, with its published numbers, in
\cref{app:published}.

\paragraph{Budgets and interaction accounting.} Every environment interaction
counts, warm-up included (\cref{tab:warmup}); \hyar{}'s reward-free
pre-training episodes are added to its step axis. \ours{} runs
$\min(\text{reference}, 5{\times}10^5)$ steps per task (\cref{tab:tasks}); the
model-free arms run the reference budgets. All arms are scored on their first
$B$ environment interactions, with $B$ our budget on the task, pre-training
included: an arm that ran longer is cut at $B$ and its final five evaluations
within the window are its score. Scoring each arm at its own, longer budget
would not be neutral: several baselines decline late in training (\hyar{} on
\HardMove{4}, for instance), so a longer budget can lower their score; those
scores are in \cref{tab:e1-full}. \dlpa{}'s planner makes the reference budgets
impractical (its decision cost is paid at every training step), so it runs to
its \emph{converged} budget: the budget of \citet[Fig.~4]{dlpa} or, where it
is not within one published standard deviation of its published return there,
until it converges, capped at ours (\cref{tab:dlpa-repro}).
Its cells are never compared at a budget it did not run.

\paragraph{Baselines and tuning.} \pdqn{}, \patdd{}, \hhqn{}, \hppo{} and
\hyar{} run from the reference code with its per-task settings, verified file
by file; for \hyar{}, \pdqn{} and \hhqn{} we run the TD3 variants, whose
published numbers are the strongest. \dlpa{} is ported from its official code
and verified bit-exact against it (\cref{app:baselines}). \chdp{} has no
released code and appears only as published numbers. Every baseline runs
with the per-task settings its authors published for these tasks, without
re-tuning; while adding our own components we tried
variants on every trained task except \HardMove{8}, mostly on \Goal{} and
\HardGoal{}, then fixed one configuration; \HardMove{8} was not used to choose
any component or setting. Our own settings are reduced from the
DreamerV3 defaults for environment scale and budget and are the same on every
task except where \cref{app:baselines:ours} states otherwise.
Our replay ratio (256 replayed steps per environment step, \cref{tab:hyper})
is also higher than the model-free baselines' defaults. Raising the replay
ratio of model-free agents is known to require additional stabilization
(ensembles, regularization or resets;
\citealp{chen2021redq,hiraoka2022droq,nikishin2022primacy,doro2023replayratio}),
and building such a variant of \hyar{} is outside the scope of this work. The
model-based baseline \dlpa{} already performs more updates per environment
step than \ours{} (25 updates of batch 64 after every step).

\paragraph{Sample efficiency and statistics.} A fixed, external threshold
(90\% of \hyar{}'s published score) is defined for every arm, including \dlpa{}
at its shorter budget; the area under the curve is not, and appears only over
our own budget (\cref{app:extra}), which also repeats the analysis at 75\% and
100\%. For ablations we report the inter-quartile mean with 95\% bootstrap
intervals~\citep{agarwal2021precipice} together with the worst seed and the
number of failing seeds, because at five seeds the tail is more informative
than a difference in means.

\subsection{Action representation}
\label{app:protocol:repr}
The world model receives the hybrid action as
$[\,\mathrm{onehot}(\modek)\;|\;\paramz_1\;|\;\dots\;|\;\paramz_\Kmodes\,]$ with
every slice except the selected mode's set to zero (\emph{blocked} layout). Each
parameter slot belongs to exactly one mode and no two modes alias the same
input unit, which matters when parameter semantics are heterogeneous. When they
are homogeneous a \emph{padded} layout $[\,\mathrm{onehot}(\modek)\;|\;\paramz_\modek\,]$
is equivalent and smaller; we use it on \HardMoveN{}. The layout is set
independently of the actor's parameter head.

\subsection{Training details}
\label{app:protocol:train}
Replay samples windows of $16{\times}32$ steps with one update every two
environment steps; windows may cross episode boundaries, where the RSSM state
is reset, and truncation is not written as termination. Imagination starts from
every posterior state of the batch, with step weights
$\prod_{t'<t}\discount\hat c_{t'}$, where $\hat c_0$ is 0 if the \emph{real}
start state is terminal and 1 otherwise: on tasks whose episodes often end
after one or two decisions (\Goal{} and \HardGoal{} early in training), a
randomly drawn start is often terminal, and without this weight much of the actor-critic
signal would be built on imagined continuations of finished episodes. The
return scale $S$ of \cref{sec:method:cme} is DreamerV3's: $\max(1,P_{95}-P_{5})$
of the $\lambda$-returns, with both percentiles averaged at decay 0.99; the
actor loss uses the same scale. \CME{} subtracts the stop-gradient baseline
$b=\sum_j\pi_j V_j$ from both of its terms, which changes neither gradient. The
continuous entropy uses the pre-squash Gaussian as a proxy for the squashed
distribution. \SWR{} uses each task's own success rule and is off on
\Platform{}, which has no sparse success event. \Epsmode{} changes only the
acting policy; imagination, the losses and evaluation are unchanged. The random
prefill follows one rule: where a random policy succeeds in fewer than one
episode in a thousand (\Goal{}, \HardGoal{}) we prefill $5\,000$ steps and use
\epsmode{}; elsewhere we prefill $1\,000$ steps without it. Prefill counts
against the budget.

\subsection{Interference without harm}
\label{app:protocol:interference}
We measure interference directly: an update driven by one mode's parameters is
applied to the actor, and the resulting change in the other modes' parameters
is divided by the change in its own (the \emph{interference ratio}; zero by
construction for \texttt{sub}). The measurement confirms that a shared head
transfers updates across modes on the heterogeneous tasks, but not that this
costs score (\cref{sec:exp:e2}). A plausible reading is that \CME{}, which
trains every candidate's parameters toward its own value, supplies each mode
with a signal strong enough to absorb the disturbance.

\section{Reproducibility}
\label{app:repro}
\paragraph{Settings.} \Cref{tab:hyper} lists our settings against the DreamerV3
defaults, with the reason for each change. The same configuration is used for
every experiment reported here; the defaults block of our configuration has every component switched
off, so every ablation row is the task's configuration with switches set back:
a leave-one-out row switches off one component of the full method, and the
ladder rows switch the components on one at a time, in a fixed order, from the
base with all of them off.
Baselines use the settings of the reference code (\cref{app:baselines}). Seeds are
0--4 for every arm, the seed loop of the reference drivers. Our agent is a PyTorch re-implementation that follows the official JAX code
of DreamerV3; every difference we know of is listed in \cref{tab:hyper}, and we
did not test numerical equivalence with the official code. It mixes two
DreamerV3 versions in two places, both stated in \cref{tab:hyper}: a pathwise
parameter gradient as in the 2023 preprint (arXiv:2301.04104v1), with the
critic's bootstrap values detached in the $\lambda$-return term (the current
version uses REINFORCE for every action; \CME{}'s parameter term needs a
pathwise path, and it does differentiate through the critic, whose weights it
does not update), and the output scale 0.01 of the current version.

\begin{table}[H]
\centering\small
\begin{tabular}{lll}
\toprule
Setting & DreamerV3 & Ours (reason) \\
\midrule
deterministic state & 512 & 256 (4--17-dim observations) \\
stochastic state & $32{\times}32$ & $16{\times}16$ (same) \\
MLP width / embed & 512 / 256 & 256 / 128 (same) \\
imagination horizon & 15 & 10 (short episodes) \\
discount $\discount$ & 0.997 & 0.99 (short episodes) \\
model / actor / critic lr & 1e-4 / 3e-5 / 3e-5 & 3e-4 / 1e-4 / 1e-4 (budgets $\le 5\cdot10^5$) \\
optimizer & Adam, $\epsilon$ 1e-8 model / 1e-5 actor-critic (2023) & Adam, $\epsilon$ 1e-8 \\
batch $\times$ length & $16{\times}64$ & $16{\times}32$ \\
replayed steps per env step & 32 (32--1024 per domain) & 256 (update every 2 steps) \\
imagination starts & all replayed states & all replayed states \\
parameter gradient & pathwise (2023) / REINFORCE (current) & pathwise, critic values detached \\
entropy coefficient & 3e-4 & 3e-4, discrete and continuous, every task \\
output-layer scale & 0.01 (current) & 0.01; parameter head 1 \\
parameter head & -- & \texttt{multi} (\HardMoveN{}), \texttt{sub} elsewhere \\
world-model action layout & -- & padded (\HardMoveN{}), blocked elsewhere \\
parameter mean bound & -- & $5\tanh(\mu/5)$ (changes $\tanh\mu$ by $<0.01$) \\
\CME{} candidates & -- & all if $\Kmodes\le16$, else top-8 + 8 random \\
\SWR{} fraction $\rho$ & -- & 0.25 (off on \Platform{}) \\
\SWR{} success event & -- & a goal (\Goal{}, \HardGoal{}); reward $>4$ (MPE) \\
\mureg{} & -- & 0.01, threshold $|\mu|=3$, every task \\
\epsmode{} & -- & 0.1$\to$0 over $5\cdot10^4$ steps (\Goal{}, \HardGoal{}) \\
prefill & -- & 5\,000 (\Goal{}/\HardGoal{}), 1\,000 elsewhere; counted \\
\bottomrule
\end{tabular}

\caption{Our settings, identical on every task except where a task is named.}
\label{tab:hyper}
\end{table}

\paragraph{Training cost of \CME{}.}
\label{app:cost}
\Cref{tab:cost} reports the time of one full update (world model, actor and
critic) for the full method, for the base configuration of the ladder
(\CME{}, \epsshort{} and \muregshort{} off; \SWR{} only changes which windows are
sampled) and for PA-Dreamer. The world-model part takes the same time in every
variant. \CME{} multiplies the update by about 1.4 with two or three
candidates and by 2 to 2.6 with eleven to sixteen; with the candidate cap it
grows little beyond $\Kmodes{=}16$ (\HardMove{8}). PA-Dreamer's actor, with one output
vector, is the cheapest. Decision time is unchanged by every training
component.
\begin{table}[H]
\centering\small
\setlength{\tabcolsep}{4pt}
\begin{tabular}{lrrrrrr}
\toprule
Task & $\Kmodes$ & candidates & base & full method & ratio & PA-Dreamer \\
\midrule
\Goal         & 3   & 3  & 115 & 169 & 1.47$\times$ & 103 \\
\HardGoal     & 11  & 11 & 122 & 247 & 2.03$\times$ & 100 \\
\CatchPoint   & 2   & 2  & 116 & 159 & 1.37$\times$ & --  \\
\HardMove{4}  & 16  & 16 & 112 & 262 & 2.34$\times$ & --  \\
\HardMove{8}  & 256 & 16 & 111 & 284 & 2.56$\times$ & --  \\
\bottomrule
\end{tabular}

\caption{Milliseconds per full update (batch $16\times32$), single process
with nothing else running, one RTX A4000. Base: \CME{}, \epsshort{} and
\muregshort{} off. Ratio: full method over base. Measured on five of the seven
tasks.}
\label{tab:cost}
\end{table}

\section{Published numbers}
\label{app:published}
\Cref{tab:published} quotes the TD3-based block and the \hppo{} column of
\citet{li2022hyar}, which our model-free arms correspond to, including \HardMove{10}, which we
do not train (\cref{app:protocol}). \Cref{tab:published-dlpa} quotes
\dlpa{}'s published episode returns (our tables report the success rate,
\cref{tab:e1}; our \dlpa{} runs in that metric: \cref{tab:dlpa-repro}) and \chdp{}'s published results, which we cannot reproduce
because no code is available; we make no claim of superiority over \chdp{}.

\begin{table}[H]
\centering\small
\begin{tabular}{lccccc}
\toprule
Task & \hppo & \patdd & \pdqn-TD3 & \hhqn-TD3 & \hyar-TD3 \\
\midrule
\Platform     & 0.80\,$\pm$\,0.02 & 0.94\,$\pm$\,0.10 & 0.93\,$\pm$\,0.03 & 0.62\,$\pm$\,0.23 & 0.98\,$\pm$\,0.01 \\
\Goal         & 0.00\,$\pm$\,0.00 & 0.00\,$\pm$\,0.00 & 0.71\,$\pm$\,0.07 & 0.00\,$\pm$\,0.00 & 0.78\,$\pm$\,0.03 \\
\HardGoal     & 0.00\,$\pm$\,0.00 & 0.44\,$\pm$\,0.05 & 0.06\,$\pm$\,0.07 & 0.01\,$\pm$\,0.01 & 0.60\,$\pm$\,0.07 \\
\CatchPoint   & 0.69\,$\pm$\,0.09 & 0.82\,$\pm$\,0.10 & 0.89\,$\pm$\,0.07 & 0.27\,$\pm$\,0.05 & 0.90\,$\pm$\,0.03 \\
\HardMove{4}  & 0.09\,$\pm$\,0.02 & 0.66\,$\pm$\,0.13 & 0.85\,$\pm$\,0.10 & 0.52\,$\pm$\,0.17 & 0.93\,$\pm$\,0.02 \\
\HardMove{6}  & 0.05\,$\pm$\,0.01 & 0.04\,$\pm$\,0.02 & 0.74\,$\pm$\,0.08 & 0.29\,$\pm$\,0.13 & 0.92\,$\pm$\,0.04 \\
\HardMove{8}  & 0.04\,$\pm$\,0.01 & 0.06\,$\pm$\,0.02 & 0.05\,$\pm$\,0.02 & 0.05\,$\pm$\,0.02 & 0.89\,$\pm$\,0.03 \\
\HardMove{10} & 0.05\,$\pm$\,0.01 & 0.07\,$\pm$\,0.02 & 0.05\,$\pm$\,0.02 & 0.05\,$\pm$\,0.02 & 0.75\,$\pm$\,0.05 \\
\bottomrule
\end{tabular}

\caption{Published success rates (return on \Platform{}), \citet{li2022hyar},
Table~2, TD3-based methods.}
\label{tab:published}
\end{table}

\begin{table}[H]
\centering\small
\setlength{\tabcolsep}{3pt}
\begin{tabular}{lcccccccc}
\toprule
 & \Platform & \Goal & \HardGoal & \CatchPoint & HM4 & HM6 & HM8 & HM10 \\
\midrule
\dlpa{} (published) & 0.92\,$\pm$\,0.05 & 28.75\,$\pm$\,6.91 & 28.38\,$\pm$\,2.88 & 7.56\,$\pm$\,4.86 & 6.29\,$\pm$\,5.74 & 8.48\,$\pm$\,5.45 & 7.80\,$\pm$\,6.27 & 6.35\,$\pm$\,9.97 \\
\chdp{} (published) & 99.7\,$\pm$\,0.2 & 80.9\,$\pm$\,4.9 & 79.5\,$\pm$\,5.0 & 93.8\,$\pm$\,0.6 & 94.2\,$\pm$\,1.7 & 93.9\,$\pm$\,1.0 & 90.6\,$\pm$\,2.2 & 79.8\,$\pm$\,5.4 \\
\bottomrule
\end{tabular}

\caption{\dlpa{}: published episode return (mean of the last 10 evaluations),
Table~1 of \citet{dlpa}. \chdp{}: success rate in percent, mean\,$\pm$\,std
over 5 runs, each the mean of its final five evaluations, Table~1 of
\citet{chdp}.}
\label{tab:published-dlpa}
\end{table}

\paragraph{Reproduction quality.}
\Cref{tab:repro} compares every model-free arm we ran with its published
TD3-block result, at the arm's own budget (the setting of the published
numbers) and on the common window of \cref{tab:e1}. Of \cReproN{} comparable
cells, \cReproIn{} lie within
two published standard deviations; \cReproLow{} lie below and \cReproHigh{}
above, so the deviations do not favor our method systematically: the largest
ones are in favor of a baseline (\hppo{} on \HardMoveN{}). For \pdqn{} on
\Platform{} and \patdd{} on \CatchPoint{}, the released code run in our
environments does not reach the published result; the other cells below
$-2\sigma$ are discussed next.

\emph{\hyar{} on \HardMove{4}/\HardMove{8}.} On the common window our runs
match the published values; they fall below only at the full $10^6$ budget,
where the success rate declines late in training (\cref{fig:curves-full}).
\citet{li2022hyar} do not define which end-of-training statistic their Table~2
reports (last evaluation, an average of the last few, or the best one), so a
late decline is invisible in the published number under some of these
definitions. We report the final-5 mean for every arm.

\emph{\patdd{} and \hhqn{} on \HardGoal{}.} The reference \HardGoal{} drivers of
these two arms, unlike their drivers for the other tasks, anneal the
exploration-noise standard deviation linearly from 1.0 to 0.1 over the first
1000 episodes; our port does the same. With the annealing, \hhqn{} reproduces
its published value. \patdd{} stays below it: four of five seeds learn (best
evaluation between 0.38 and 0.67), and three of them later return to the
shoot-at-once optimum. The remaining known difference is the episode cap: the
published \Goal{}/\HardGoal{} values of both arms were obtained with 250 steps,
where we use 150 for every arm (\cref{app:baselines}).

\emph{\daac{}.} \citet{daac2026} report hybrid-action results on the same
suite against \patdd{}, \pdqn{} and \hhqn{} only. At its reported budgets its
learning curves stay below \hyar{}'s published TD3 scores on \Platform{},
\Goal{}, \HardGoal{}, \HardMove{6} and \HardMove{8}, and above them on
\CatchPoint{} and \HardMove{4}, where it may be a stronger model-free baseline
than \hyar{}. We do not run it.

\emph{\pdqn{} on \HardMove{4}/\HardMove{6}.} The audit found no difference from
the reference driver: arguments, Ornstein--Uhlenbeck noise, $\epsilon$ schedule
and evaluation all match.
\begin{table}[H]
\centering\small
\setlength{\tabcolsep}{5pt}
\begin{tabular}{llcccr}
\toprule
Arm & Task & published & ours, own budget & ours, common window & $\Delta/\sigma_{\mathrm{pub}}$ \\
\midrule
\hyar & \Platform & 0.98\,$\pm$\,0.01 & 0.97\,$\pm$\,0.03 & 0.98\,$\pm$\,0.01 & $-$0.8 \\
 & \Goal & 0.78\,$\pm$\,0.03 & 0.74\,$\pm$\,0.03 & 0.74\,$\pm$\,0.02 & $-$1.3 \\
 & \HardGoal & 0.60\,$\pm$\,0.07 & 0.57\,$\pm$\,0.11 & 0.59\,$\pm$\,0.08 & $-$0.4 \\
 & \CatchPoint & 0.90\,$\pm$\,0.03 & 0.85\,$\pm$\,0.15 & 0.70\,$\pm$\,0.18 & $-$1.7 \\
 & \HardMove{4} & 0.93\,$\pm$\,0.02 & 0.84\,$\pm$\,0.09 & 0.95\,$\pm$\,0.01 & \textbf{$-$4.6} \\
 & \HardMove{6} & 0.92\,$\pm$\,0.04 & 0.87\,$\pm$\,0.06 & 0.93\,$\pm$\,0.04 & $-$1.2 \\
 & \HardMove{8} & 0.89\,$\pm$\,0.03 & 0.81\,$\pm$\,0.13 & 0.89\,$\pm$\,0.06 & \textbf{$-$2.7} \\
\midrule
\pdqn & \Platform & 0.93\,$\pm$\,0.03 & 0.84\,$\pm$\,0.13 & 0.84\,$\pm$\,0.13 & \textbf{$-$2.9} \\
 & \Goal & 0.71\,$\pm$\,0.07 & 0.61\,$\pm$\,0.20 & 0.61\,$\pm$\,0.20 & $-$1.4 \\
 & \HardGoal & 0.06\,$\pm$\,0.07 & 0.07\,$\pm$\,0.08 & 0.07\,$\pm$\,0.08 & $+$0.1 \\
 & \CatchPoint & 0.89\,$\pm$\,0.07 & 0.92\,$\pm$\,0.03 & 0.92\,$\pm$\,0.05 & $+$0.4 \\
 & \HardMove{4} & 0.85\,$\pm$\,0.10 & 0.43\,$\pm$\,0.07 & 0.30\,$\pm$\,0.05 & \textbf{$-$4.2} \\
 & \HardMove{6} & 0.74\,$\pm$\,0.08 & 0.33\,$\pm$\,0.09 & 0.34\,$\pm$\,0.06 & \textbf{$-$5.2} \\
 & \HardMove{8} & 0.05\,$\pm$\,0.02 & 0.14\,$\pm$\,0.07 & 0.14\,$\pm$\,0.05 & \textbf{$+$4.7} \\
\midrule
\hhqn & \Platform & 0.62\,$\pm$\,0.23 & 0.42\,$\pm$\,0.22 & 0.42\,$\pm$\,0.22 & $-$0.9 \\
 & \Goal & 0.00\,$\pm$\,0.00 & 0.00\,$\pm$\,0.00 & 0.00\,$\pm$\,0.00 & $+$0.0 \\
 & \HardGoal & 0.01\,$\pm$\,0.01 & 0.02\,$\pm$\,0.04 & 0.02\,$\pm$\,0.04 & $+$1.0 \\
 & \CatchPoint & 0.27\,$\pm$\,0.05 & 0.24\,$\pm$\,0.14 & 0.33\,$\pm$\,0.17 & $-$0.7 \\
 & \HardMove{4} & 0.52\,$\pm$\,0.17 & 0.36\,$\pm$\,0.19 & 0.55\,$\pm$\,0.19 & $-$0.9 \\
 & \HardMove{6} & 0.29\,$\pm$\,0.13 & 0.29\,$\pm$\,0.17 & 0.17\,$\pm$\,0.05 & $+$0.0 \\
 & \HardMove{8} & 0.05\,$\pm$\,0.02 & 0.04\,$\pm$\,0.02 & 0.05\,$\pm$\,0.04 & $-$0.4 \\
\midrule
\patdd & \Platform & 0.94\,$\pm$\,0.10 & 0.92\,$\pm$\,0.03 & 0.92\,$\pm$\,0.03 & $-$0.2 \\
 & \Goal & 0.00\,$\pm$\,0.00 & 0.00\,$\pm$\,0.00 & 0.00\,$\pm$\,0.00 & $+$0.0 \\
 & \HardGoal & 0.44\,$\pm$\,0.05 & 0.14\,$\pm$\,0.18 & 0.14\,$\pm$\,0.18 & \textbf{$-$5.9} \\
 & \CatchPoint & 0.82\,$\pm$\,0.10 & 0.50\,$\pm$\,0.05 & 0.59\,$\pm$\,0.17 & \textbf{$-$3.2} \\
 & \HardMove{4} & 0.66\,$\pm$\,0.13 & 0.85\,$\pm$\,0.07 & 0.49\,$\pm$\,0.18 & $+$1.4 \\
 & \HardMove{6} & 0.04\,$\pm$\,0.02 & 0.04\,$\pm$\,0.02 & 0.07\,$\pm$\,0.04 & $-$0.2 \\
 & \HardMove{8} & 0.06\,$\pm$\,0.02 & 0.08\,$\pm$\,0.03 & 0.08\,$\pm$\,0.03 & $+$1.1 \\
\midrule
\hppo & \Platform & 0.80\,$\pm$\,0.02 & 0.81\,$\pm$\,0.03 & 0.81\,$\pm$\,0.03 & $+$0.5 \\
 & \Goal & 0.00\,$\pm$\,0.00 & 0.00\,$\pm$\,0.00 & 0.00\,$\pm$\,0.00 & $+$0.1 \\
 & \HardGoal & 0.00\,$\pm$\,0.00 & 0.00\,$\pm$\,0.00 & 0.00\,$\pm$\,0.00 & $+$0.0 \\
 & \CatchPoint & 0.69\,$\pm$\,0.09 & 0.76\,$\pm$\,0.06 & 0.72\,$\pm$\,0.06 & $+$0.8 \\
 & \HardMove{4} & 0.09\,$\pm$\,0.02 & 0.64\,$\pm$\,0.07 & 0.64\,$\pm$\,0.12 & \textbf{$+$27.6} \\
 & \HardMove{6} & 0.05\,$\pm$\,0.01 & 0.63\,$\pm$\,0.10 & 0.59\,$\pm$\,0.19 & \textbf{$+$57.8} \\
 & \HardMove{8} & 0.04\,$\pm$\,0.01 & 0.52\,$\pm$\,0.16 & 0.14\,$\pm$\,0.05 & \textbf{$+$47.6} \\
\bottomrule
\end{tabular}

\caption{Reproduced against published success rates (return on \Platform{};
\citet{li2022hyar}, Table~2, TD3 block). Own budget: the arm's reference budget,
as in the publication. Common window: \cref{tab:e1}. Last column: deviation of
the own-budget mean in units of the published standard deviation (floor 0.01);
bold beyond $\pm2$.}
\label{tab:repro}
\end{table}

\emph{\dlpa{}.} \Cref{tab:dlpa-repro} compares our \dlpa{} runs with its
published results in its own metric, the episode return averaged over the last
10 evaluations, under our protocol (\cref{app:baselines}). At the budget of
\citet[Fig.~4]{dlpa}, our runs are within one published standard deviation of
the published return on \CatchPoint{}, \HardMove{4} and \HardMove{6}, and
below it by more on \Platform{}, \Goal{}, \HardGoal{} and \HardMove{8}; by the
budget rule (\cref{app:protocol}) these four runs are extended until they
converge. At the
converged budget every task is within one published standard deviation or
above, except \HardGoal{}: two of its five seeds stay at the shoot-at-once
optimum (success rate 0) and are kept, as is the one such seed on \Goal{}
(\cref{app:baselines}).
\begin{table}[H]
\centering\small
\setlength{\tabcolsep}{4pt}
\begin{tabular}{lcrcrrcr}
\toprule
 & & \multicolumn{3}{c}{budget of \citet[Fig.~4]{dlpa}} & \multicolumn{3}{c}{converged budget} \\
\cmidrule(lr){3-5}\cmidrule(l){6-8}
Task & published & steps & ours & $z$ & steps & ours & $z$ \\
\midrule
\Platform & 0.92\,$\pm$\,0.05 & 2e4 & 0.85\,$\pm$\,0.28 & $-$1.5 & 5e4 & 0.98\,$\pm$\,0.02 & +1.1 \\
\Goal & 28.75\,$\pm$\,6.91 & 2e4 & 16.54\,$\pm$\,11.51 & $-$1.8 & 1.5e5 & 25.99\,$\pm$\,16.09 & $-$0.4 \\
\HardGoal & 28.38\,$\pm$\,2.88 & 5e4 & 11.66\,$\pm$\,14.68 & \textbf{$-$5.8} & 9e4 & 13.17\,$\pm$\,15.82 & \textbf{$-$5.3} \\
\CatchPoint & 7.56\,$\pm$\,4.86 & 2e5 & 6.90\,$\pm$\,0.64 & $-$0.1 & \multicolumn{3}{c}{not extended} \\
\HardMove{4} & 6.29\,$\pm$\,5.74 & 1.5e5 & 6.91\,$\pm$\,0.49 & +0.1 & \multicolumn{3}{c}{not extended} \\
\HardMove{6} & 8.48\,$\pm$\,5.45 & 8e4 & 5.69\,$\pm$\,0.80 & $-$0.5 & \multicolumn{3}{c}{not extended} \\
\HardMove{8} & 7.80\,$\pm$\,6.27 & 2e4 & $-$17.44\,$\pm$\,1.83 & \textbf{$-$4.0} & 1.5e5 & 5.43\,$\pm$\,0.20 & $-$0.4 \\
\bottomrule
\end{tabular}

\caption{\dlpa{}: published episode return \citep[Table~1]{dlpa} against our
runs, per seed the mean of the last 10 evaluations, mean\,$\pm$\,std over 5
seeds. $z$: deviation from the published mean in units of the published
standard deviation; bold beyond $\pm2$. Converged budget: the budget of
\cref{tab:e1}.}
\label{tab:dlpa-repro}
\end{table}

\section{Environments}
\label{app:env}

\subsection{Tasks and protocol constants}
\begin{table}[H]
\centering\small
\setlength{\tabcolsep}{4pt}
\begin{tabular}{lrrrrlrrrr}
\toprule
 & & & & & & \multicolumn{4}{c}{budget (env steps)} \\
\cmidrule(l){7-10}
Task & $\Kmodes$ & $\sum_\modek \paramdim{\modek}$ & eval every & max steps & metric & reference & ours & \dlpa{} & \dlpa{} (published) \\
\midrule
\Platform & 3 & 3 & 500 & 250 & return & 2e5 & 2e5 & 5e4 & 2e4 \\
\Goal & 3 & 4 & 500 & 150 & success & 3e5 & 3e5 & 1.5e5 & 2e4 \\
\HardGoal & 11 & 12 & 500 & 150 & success & 3e5 & 3e5 & 9e4 & 5e4 \\
\CatchPoint & 2 & 2 & 2\,500 & 25 & success & 1e6 & 5e5 & 2e5 & 2e5 \\
\HardMove{4} & 16 & 16 & 2\,500 & 30 & success & 1e6 & 5e5 & 1.5e5 & 1.5e5 \\
\HardMove{6} & 64 & 64 & 2\,500 & 30 & success & 1e6 & 5e5 & 8e4 & 8e4 \\
\HardMove{8} & 256 & 256 & 2\,500 & 30 & success & 1e6 & 5e5 & 1.5e5 & 2e4 \\
\HardMove{10} & 1024 & 1024 & -- & -- & (latency only) & -- & -- & -- & -- \\
\bottomrule
\end{tabular}

\caption{Tasks and the protocol constants of the reference implementation.
Last four columns: budgets in environment steps.}
\label{tab:tasks}
\end{table}
Success is defined as in the reference implementation: on \Goal{} and
\HardGoal{} an episode succeeds when its terminal step yields the goal reward
of 50 (an episode that ends without a goal ends with the negative distance of
the ball to the goal); on
\CatchPoint{} and \HardMoveN{} when some step yields a reward above 4.
\Platform{} reports the normalized episode return.

\subsection{Notes on two reference implementations}
\label{app:env:packing}
\label{app:env:hhqn}
\emph{\CatchPoint{} parameters.} The reference drivers give the two
\CatchPoint{} modes one shared scalar parameter, the movement angle, which the
catch mode does not use. Following the PAMDP definition, we give each mode its
own parameter; our environment adapter maps it to the same executed action,
and a golden-trace test against reference rollouts confirms that the
environment is unchanged. The only difference is that every agent's parameter
vector has two entries instead of one.

\emph{\hhqn{}-TD3.} The released implementation computes its second critic
head through the first head's layers, so clipped double-Q is absent; it has no
target policy smoothing and no delayed policy update, and it does not update
its target parameter network. We ran the file unmodified, as published. These
properties are consistent with its wide seed variance (published
$0.62\pm0.23$ on \Platform{}).

\subsection{Degeneracy of \HardMoveN{}}
\label{app:env:degeneracy}
\begin{table}[H]
\centering\small
\begin{tabular}{lrrr}
\toprule
Task & nominal modes & distinct motions & zero-motion modes \\
\midrule
\HardMove{4}  & 16   & 9   & 4  \\
\HardMove{6}  & 64   & 19  & 10 \\
\HardMove{8}  & 256  & 81  & 16 \\
\HardMove{10} & 1024 & 211 & 34 \\
\bottomrule
\end{tabular}

\caption{Opposing actuators cancel, so the nominal modes of \HardMoveN{}
collapse to far fewer distinct motions.}
\label{tab:degeneracy}
\end{table}
Mode $\modek$ of \HardMoveN{} applies the sum of the actuator directions encoded
by the bits of $\modek$, so opposing actuators cancel: the $2^n$ nominal modes
collapse to 9, 19, 81 and 211 distinct motions for $n=4,6,8,10$, of which 4, 10,
16 and 34 produce no motion at all. A learned policy leaves these modes: the
share of dead modes it selects falls from its uniform value of \cDeadFourUni\%,
\cDeadSixUni\% and \cDeadEightUni\% for $n=4,6,8$ to \cDeadFour\%, \cDeadSix\%
and \cDeadEight\% in the last tenth of training (mean over seeds). The released
planner loops over the nominal modes serially (\cref{app:baselines}), so its
cost (\cref{tab:e6}) grows with their number, not with the number of distinct
behaviors.

\section{Baseline reproduction notes}
\label{app:baselines}
\paragraph{Sources.} \pdqn{}, \patdd{}, \hhqn{}, \hppo{} and \hyar{}: agent files
vendored unchanged from the reference repository, driver logic ported, every
per-task setting checked against the drivers' defaults. \dlpa{}: ported from the
official code and compared bit-exact on the CPU (weights, forward pass, losses
and the optimizer step, planner, replay windows) against it.

\paragraph{HyAR's pre-training.} \hyar{} trains its action
representation on reward-free random episodes before its first update. These
environment steps are not counted by the reference code; we add them to its
step axis in every curve, in \cref{tab:efficiency} and in \cref{fig:pareto};
\cref{tab:efficiency} also reports its steps without them.

\paragraph{Warm-up and interaction accounting.} \Cref{tab:warmup} lists
every arm's warm-up phase and whether it is counted.
\begin{table}[H]
\centering\small
\begin{tabular}{lp{7.6cm}p{4.2cm}}
\toprule
Arm & Warm-up & Counted in budget \\
\midrule
\hyar     & pre-training: 5\,000 random episodes (20\,000 on \Goal{}/\HardGoal{}) that train only the action VAE; then 128 random steps before the first policy update & yes, both (the reference code does not count the pre-training; we add it) \\
\dlpa     & first 50 episodes with random actions, no model update & yes \\
\patdd, \hhqn & 128 steps with the noisy (not random) policy, no update & yes \\
\pdqn     & 128 steps (500 on \Platform{}), $\epsilon$-greedy from $\epsilon{=}1$, no update & yes \\
\hppo     & none (on-policy) & -- \\
\ours     & random prefill: 5\,000 steps on \Goal{}/\HardGoal{}, 1\,000 elsewhere; then \epsmode{} ($\varepsilon$ 0.1$\to$0 over $5\cdot10^4$ steps) on \Goal{}/\HardGoal{} only & yes \\
\bottomrule
\end{tabular}

\caption{Warm-up phases. Every counted interaction is on the x-axis of every
figure; \hyar{}'s pre-training is added to its axis.}
\label{tab:warmup}
\end{table}

\paragraph{Deviations from the reference code.}
(i) Evaluation runs in a separate environment under a preserved random stream;
the reference evaluates inside the training loop, which changes the training
environment's random state.
(ii) One episode cap per task for every arm: 25 on \CatchPoint{}, where
\hppo{}'s driver uses 30, and 150 on \Goal{} and \HardGoal{}, the value of
\hyar{}'s drivers, where \patdd{} and \hhqn{} were published with 250.
(iii) A \Goal{} budget of $3\cdot10^5$ for every model-free arm, where
\hppo{}'s driver uses $2\cdot10^5$.
(iv) \hyar{}'s zero-padding of unused parameter slots follows its \HardGoal{}
driver on \Goal{} as well, where its \Goal{} loop omits it.
(v) \CatchPoint{} gives each mode its own parameter (\cref{app:env:packing});
\hyar{} keeps the reference embedding sizes (4/4).
(vi) Reference arguments that no driver reads (\pdqn{}'s \texttt{multipass},
\texttt{split}, \texttt{action-input-layer}, \texttt{reward-scale}) are not
reproduced, and where \hyar{}'s paper and its drivers disagree (episode
horizons in its Appendix~B.1) we follow the drivers. All deviations will be
documented with the code release.

\paragraph{Reference settings that are easy to miss.} The following are
reproduced from the reference drivers, not changed. \pdqn{}'s hyperparameters
differ per task (discount, warm-up, buffer size, target rates, learning rate and
gradient clipping); its hand-coded passthrough initialization is active on
\Platform{} and \Goal{} only. \hyar{}'s replay buffers differ per task ($10^6$
transitions on \HardMoveN{}, $10^5$ elsewhere; $2\cdot10^6$ for the embedding
buffer on \Goal{} and \HardGoal{}), and its target-policy noise is 0.1 where the
\patdd{} and \hhqn{} drivers use 0.2. \patdd{} and \hhqn{} anneal their
exploration noise from 1.0 to 0.1 over the first 1000 episodes on \HardGoal{}
only; elsewhere it is a constant 0.1.

\paragraph{\dlpa{}.} (i) The released code augments the \CatchPoint{}
observation with two history features (a failed-catch counter and a step counter)
that the paper does not describe and no other method receives; we run without
them, so all methods observe the same state. (ii) The released harness trains on
25-step episodes on every task and cuts the training episode at each evaluation;
we use the protocol caps and do neither. (iii) Where the paper's and the code's
hyperparameters differ we use the code's. (iv) Budgets: the budget of \citet[Fig.~4]{dlpa}; where our run is not within one
published standard deviation of the published return there, it is extended
until it converges, capped at our budget
(\cref{tab:tasks}); convergence is a last-quarter slope within evaluation
noise. \Cref{tab:dlpa-repro} compares both budgets with the published returns.
(v) \HardGoal{} is not runnable from the released code (a missing padding
function); our environment adapter provides it.
(vi) \HardMove{8}: at that budget our reproduction is well below the
published return, and the official 25-step episode cap does not close the gap
(two seeds), so the run is extended by the rule above (\cref{tab:tasks}). At
the extended budget it reaches the best mean of every arm on \HardMove{8}
(\cref{tab:e1}). (vii) On \Goal{} and \HardGoal{} some seeds lock into
the shoot-at-once optimum (episode length 1, return $-6.2$) from the start;
they are kept (no seed selection). Extending these seeds beyond the converged
budget (to $3\cdot10^5$ steps on \Goal{} and $1.6\cdot10^5$ on \HardGoal{}) does
not free them, so a longer budget would not change \dlpa{}'s cells. (viii) Latency: the released planner loops
over modes serially and synchronizes with the CPU at every decision; we keep
this structure, which is why a GPU helps \dlpa{} only at small $\Kmodes$. On \Platform{},
the slowest task at small $\Kmodes$, the longest planning horizon of the suite
($L{=}10$, against 5 on \CatchPoint{} and Hard Move) dominates instead.
(ix) We compare against the strongest published form of each approach, since
our aim is an agent that stays at state-of-the-art final performance while
being sample-efficient and fast at decision time. For model-based PAMDP methods
this is \dlpa{} with planning: \citet[Sec.~6.2, Fig.~5a]{dlpa} evaluate a
Dyna-style variant that trains a policy on model rollouts and report that it
falls behind planning on \Platform{} and \Goal{}. A weaker variant that acts
faster would add a point to the trade-off of \cref{fig:pareto}, not a stronger
competitor; an arm that does not reach the threshold appears there only as a
bound. A lower planner budget shortens \dlpa{}'s decision roughly in
proportion to $I$ (\HardMove{4}, one CPU thread: 15.7\,ms at $I{=}2$,
7.9\,ms at $I{=}1$). By this scaling the released planner would act above
50\,Hz at $I{=}2$ on some tasks (\HardMove{4}) but not on \Platform{} or
\HardMove{8}, and at $I{=}1$ still not on \HardMove{8}. Whether \dlpa{} keeps
its sample efficiency at such budgets is not tested here.

\subsection{Our own settings}
\label{app:baselines:ours}
Every change from the DreamerV3 defaults is listed with its reason in
\cref{tab:hyper}. None was searched per task. The per-task choices follow fixed rules: the
parameter head (\texttt{multi} where every mode has the same single
parameter, \cref{sec:method:actor}), the world-model action layout
(\emph{padded} under the same condition, \emph{blocked} otherwise,
\cref{app:protocol:repr}), the \SWR{} success event by each task's success rule
(\SWR{} off on \Platform{}, which has no sparse success event), and
\epsshort{} with a longer prefill where a random policy almost never succeeds
(\cref{app:protocol:train}).

\section{Additional results}
\label{app:extra}

\subsection{Scores at each arm's full budget}
\Cref{tab:e1-full} scores every arm at its own budget instead of the common
window of \cref{tab:e1}: \ours{} at $\min(\text{reference}, 5\cdot10^5)$, the
model-free arms at the reference budgets (up to $10^6$ steps), \dlpa{} at its
converged budget.
\begin{table}[H]
\centering\small
\setlength{\tabcolsep}{3pt}
\begin{tabular}{lcccccccc}
\toprule
Task & Budget & \ours{} (ours) & \dlpa$^\dagger$ & \hyar & \pdqn$^\ddagger$ & \hhqn & \patdd & \hppo \\
\midrule
\Platform & 2e5 & \textbf{1.00\,$\pm$\,0.00} & 0.97\,$\pm$\,0.03 & 0.97\,$\pm$\,0.03 & 0.84\,$\pm$\,0.13 & 0.42\,$\pm$\,0.22 & 0.92\,$\pm$\,0.03 & 0.81\,$\pm$\,0.03 \\
\Goal & 3e5 & \textbf{0.77\,$\pm$\,0.03} & 0.58\,$\pm$\,0.29 & 0.74\,$\pm$\,0.03 & 0.61\,$\pm$\,0.20 & 0.00\,$\pm$\,0.00 & 0.00\,$\pm$\,0.00 & 0.00\,$\pm$\,0.00 \\
\HardGoal & 3e5 & \textbf{0.79\,$\pm$\,0.04} & 0.34\,$\pm$\,0.28 & 0.57\,$\pm$\,0.11 & 0.07\,$\pm$\,0.08 & 0.02\,$\pm$\,0.04 & 0.14\,$\pm$\,0.18 & 0.00\,$\pm$\,0.00 \\
\CatchPoint & 5e5 / 1e6 & \textbf{0.94\,$\pm$\,0.00} & 0.92\,$\pm$\,0.03 & 0.85\,$\pm$\,0.15 & 0.92\,$\pm$\,0.03 & 0.24\,$\pm$\,0.14 & 0.50\,$\pm$\,0.05 & 0.76\,$\pm$\,0.06 \\
\HardMove{4} & 5e5 / 1e6 & 0.91\,$\pm$\,0.05 & \textbf{0.94\,$\pm$\,0.01} & 0.84\,$\pm$\,0.09 & 0.43\,$\pm$\,0.07 & 0.36\,$\pm$\,0.19 & 0.85\,$\pm$\,0.07 & 0.64\,$\pm$\,0.07 \\
\HardMove{6} & 5e5 / 1e6 & \textbf{0.95\,$\pm$\,0.03} & \textit{0.94\,$\pm$\,0.01} & 0.87\,$\pm$\,0.06 & 0.33\,$\pm$\,0.09 & 0.29\,$\pm$\,0.17 & 0.04\,$\pm$\,0.02 & 0.63\,$\pm$\,0.10 \\
\HardMove{8} & 5e5 / 1e6 & 0.91\,$\pm$\,0.06 & \textbf{0.93\,$\pm$\,0.01} & 0.81\,$\pm$\,0.13 & 0.14\,$\pm$\,0.07 & 0.04\,$\pm$\,0.02 & 0.08\,$\pm$\,0.03 & 0.52\,$\pm$\,0.16 \\
\bottomrule
\end{tabular}

\caption{Final performance with every arm scored at its own budget (mean of the
final five evaluations, mean\,$\pm$\,std over 5 seeds). Budget: ours /
model-free arms. $^\dagger$At its converged, shorter budget. $^\ddagger$With a
hand-coded parameter prior. Bold: best mean; italic: mean at least the best mean minus the best arm's standard deviation, both computed before rounding.}
\label{tab:e1-full}
\end{table}

\subsection{Robustness of the score}
\Cref{tab:late} repeats \cref{tab:e1} with the mean over the last
$\min(5\cdot10^4, 25\%\text{ of the budget})$ steps of each run instead of the
final five evaluations. It is applied identically to every arm; the final-five
score remains the primary metric (\cref{sec:exp:protocol}).
\begin{table}[H]
\centering\small
\setlength{\tabcolsep}{3pt}
\begin{tabular}{lccccccc}
\toprule
Task & \ours{} (ours) & \dlpa$^\dagger$ & \hyar & \pdqn$^\ddagger$ & \hhqn & \patdd & \hppo \\
\midrule
\Platform & \textbf{1.00\,$\pm$\,0.00} & 0.98\,$\pm$\,0.03 & 0.97\,$\pm$\,0.01 & 0.89\,$\pm$\,0.03 & 0.44\,$\pm$\,0.23 & 0.91\,$\pm$\,0.08 & 0.80\,$\pm$\,0.01 \\
\Goal & \textbf{0.76\,$\pm$\,0.03} & 0.57\,$\pm$\,0.29 & 0.72\,$\pm$\,0.02 & 0.61\,$\pm$\,0.18 & 0.00\,$\pm$\,0.00 & 0.00\,$\pm$\,0.00 & 0.00\,$\pm$\,0.00 \\
\HardGoal & \textbf{0.78\,$\pm$\,0.03} & 0.34\,$\pm$\,0.28 & 0.58\,$\pm$\,0.08 & 0.06\,$\pm$\,0.08 & 0.01\,$\pm$\,0.02 & 0.18\,$\pm$\,0.17 & 0.00\,$\pm$\,0.00 \\
\CatchPoint & \textit{0.92\,$\pm$\,0.03} & \textit{0.92\,$\pm$\,0.01} & 0.72\,$\pm$\,0.17 & \textbf{0.92\,$\pm$\,0.05} & 0.31\,$\pm$\,0.14 & 0.61\,$\pm$\,0.14 & 0.70\,$\pm$\,0.04 \\
\HardMove{4} & 0.92\,$\pm$\,0.02 & \textit{0.95\,$\pm$\,0.00} & \textbf{0.95\,$\pm$\,0.01} & 0.35\,$\pm$\,0.05 & 0.63\,$\pm$\,0.13 & 0.48\,$\pm$\,0.21 & 0.65\,$\pm$\,0.12 \\
\HardMove{6} & \textbf{0.95\,$\pm$\,0.02} & \textit{0.94\,$\pm$\,0.01} & 0.93\,$\pm$\,0.03 & 0.33\,$\pm$\,0.06 & 0.18\,$\pm$\,0.07 & 0.06\,$\pm$\,0.03 & 0.57\,$\pm$\,0.20 \\
\HardMove{8} & \textbf{0.92\,$\pm$\,0.02} & \textit{0.92\,$\pm$\,0.00} & 0.86\,$\pm$\,0.09 & 0.13\,$\pm$\,0.03 & 0.05\,$\pm$\,0.03 & 0.08\,$\pm$\,0.02 & 0.15\,$\pm$\,0.05 \\
\bottomrule
\end{tabular}

\caption{Late-window mean (mean\,$\pm$\,std over 5 seeds). Bold and italic as
in \cref{tab:e1}.}
\label{tab:late}
\end{table}

\begin{table}[H]
\centering\small
\setlength{\tabcolsep}{3pt}
\begin{tabular}{lccccccc}
\toprule
Task & \ours{} (ours) & \dlpa$^\dagger$ & \hyar & \pdqn$^\ddagger$ & \hhqn & \patdd & \hppo \\
\midrule
\Platform & \textbf{0.98\,$\pm$\,0.00} & 0.96\,$\pm$\,0.06 & 0.90\,$\pm$\,0.02 & 0.87\,$\pm$\,0.05 & 0.41\,$\pm$\,0.10 & 0.88\,$\pm$\,0.06 & 0.65\,$\pm$\,0.02 \\
\Goal & \textbf{0.65\,$\pm$\,0.02} & 0.54\,$\pm$\,0.27 & 0.59\,$\pm$\,0.04 & 0.59\,$\pm$\,0.15 & 0.00\,$\pm$\,0.00 & 0.00\,$\pm$\,0.00 & 0.00\,$\pm$\,0.00 \\
\HardGoal & \textbf{0.69\,$\pm$\,0.02} & 0.33\,$\pm$\,0.27 & 0.46\,$\pm$\,0.11 & 0.02\,$\pm$\,0.03 & 0.02\,$\pm$\,0.01 & 0.22\,$\pm$\,0.15 & 0.00\,$\pm$\,0.00 \\
\CatchPoint & \textbf{0.90\,$\pm$\,0.02} & 0.92\,$\pm$\,0.03 & 0.64\,$\pm$\,0.12 & 0.84\,$\pm$\,0.03 & 0.32\,$\pm$\,0.05 & 0.39\,$\pm$\,0.13 & 0.45\,$\pm$\,0.02 \\
\HardMove{4} & \textit{0.87\,$\pm$\,0.02} & 0.94\,$\pm$\,0.02 & \textbf{0.88\,$\pm$\,0.01} & 0.33\,$\pm$\,0.03 & 0.70\,$\pm$\,0.08 & 0.24\,$\pm$\,0.10 & 0.44\,$\pm$\,0.08 \\
\HardMove{6} & \textbf{0.89\,$\pm$\,0.03} & 0.93\,$\pm$\,0.02 & 0.83\,$\pm$\,0.03 & 0.27\,$\pm$\,0.03 & 0.22\,$\pm$\,0.08 & 0.05\,$\pm$\,0.01 & 0.29\,$\pm$\,0.09 \\
\HardMove{8} & \textbf{0.84\,$\pm$\,0.02} & 0.88\,$\pm$\,0.02 & 0.70\,$\pm$\,0.04 & 0.08\,$\pm$\,0.01 & 0.06\,$\pm$\,0.02 & 0.07\,$\pm$\,0.01 & 0.07\,$\pm$\,0.01 \\
\bottomrule
\end{tabular}

\caption{Area under the evaluation curve over our budget on the task, divided by
its length (mean\,$\pm$\,std over 5 seeds). Every arm is integrated over the
same range. \dlpa{}'s runs end earlier; its curve is held at its last value to
the end of the range, so its column is not comparable, is not ranked and is
shown for completeness only. Held at its last value, \dlpa{} would have the
highest value on \CatchPoint{}, \HardMove{4}, \HardMove{6} and \HardMove{8}. \hyar{}'s curve is integrated on its own step
axis, without its pre-training. Bold and italic as in \cref{tab:e1}.}
\label{tab:auc}
\end{table}

\subsection{Sensitivity to the threshold}
\begin{table}[H]
\centering\small
\setlength{\tabcolsep}{4pt}
\begin{tabular}{llccccccc}
\toprule
Task & thr. & \ours{} (ours) & \dlpa$^\dagger$ & \hyar & \pdqn$^\ddagger$ & \hhqn & \patdd & \hppo \\
\midrule
\Platform & 75\% & 4k & \textbf{1k} & 19.5k & 5k & $>$200k\,\scriptsize(2/5) & 17k & 93k \\
 & 90\% & 4.5k & \textbf{1.5k} & 25k & 9k & $>$200k\,\scriptsize(2/5) & 21.5k & $>$200k\,\scriptsize(0/5) \\
 & 100\% & \textbf{5k} & 8.5k & 40k & 47k & $>$200k\,\scriptsize(0/5) & 27k & $>$200k\,\scriptsize(0/5) \\
\addlinespace
\Goal & 75\% & \textbf{32.5k} & 20.5k\,\scriptsize(4/5) & 60.3k & 20.5k\,\scriptsize(4/5) & $>$300k\,\scriptsize(0/5) & $>$300k\,\scriptsize(0/5) & $>$300k\,\scriptsize(0/5) \\
 & 90\% & \textbf{45k} & 47k\,\scriptsize(4/5) & 85.8k & 31.5k\,\scriptsize(4/5) & $>$300k\,\scriptsize(0/5) & $>$300k\,\scriptsize(0/5) & $>$300k\,\scriptsize(0/5) \\
 & 100\% & 105k & 73k\,\scriptsize(4/5) & \textbf{95.8k} & 78.5k\,\scriptsize(4/5) & $>$300k\,\scriptsize(0/5) & $>$300k\,\scriptsize(0/5) & $>$300k\,\scriptsize(0/5) \\
\addlinespace
\HardGoal & 75\% & \textbf{22k} & 16.5k\,\scriptsize(3/5) & 57.2k & $>$300k\,\scriptsize(0/5) & $>$300k\,\scriptsize(0/5) & $>$300k\,\scriptsize(2/5) & $>$300k\,\scriptsize(0/5) \\
 & 90\% & \textbf{29.5k} & 29k\,\scriptsize(3/5) & 105k & $>$300k\,\scriptsize(0/5) & $>$300k\,\scriptsize(0/5) & $>$300k\,\scriptsize(2/5) & $>$300k\,\scriptsize(0/5) \\
 & 100\% & \textbf{29.5k} & 29k\,\scriptsize(3/5) & 175k\,\scriptsize(4/5) & $>$300k\,\scriptsize(0/5) & $>$300k\,\scriptsize(0/5) & $>$300k\,\scriptsize(2/5) & $>$300k\,\scriptsize(0/5) \\
\addlinespace
\CatchPoint & 75\% & 15k & \textbf{7.5k} & 174k & 57.5k & $>$500k\,\scriptsize(0/5) & 338k\,\scriptsize(4/5) & 282k \\
 & 90\% & 20k & \textbf{10k} & 221k & 105k & $>$500k\,\scriptsize(0/5) & $>$500k\,\scriptsize(2/5) & 432k\,\scriptsize(4/5) \\
 & 100\% & 30k & \textbf{17.5k} & 256k\,\scriptsize(4/5) & 135k & $>$500k\,\scriptsize(0/5) & $>$500k\,\scriptsize(0/5) & $>$500k\,\scriptsize(0/5) \\
\addlinespace
\HardMove{4} & 75\% & 15k & \textbf{5k} & 185k & 122k & 32.5k & $>$500k\,\scriptsize(1/5) & 268k\,\scriptsize(3/5) \\
 & 90\% & 27.5k & \textbf{15k} & 185k & $>$500k\,\scriptsize(2/5) & 55k & $>$500k\,\scriptsize(1/5) & $>$500k\,\scriptsize(2/5) \\
 & 100\% & 40k & \textbf{22.5k} & 228k & $>$500k\,\scriptsize(0/5) & 75k\,\scriptsize(4/5) & $>$500k\,\scriptsize(0/5) & $>$500k\,\scriptsize(0/5) \\
\addlinespace
\HardMove{6} & 75\% & 25k & \textbf{15k} & 197k & 228k & 302k\,\scriptsize(3/5) & $>$500k\,\scriptsize(0/5) & 415k\,\scriptsize(3/5) \\
 & 90\% & 35k & \textbf{22.5k} & 212k & 340k\,\scriptsize(4/5) & $>$500k\,\scriptsize(0/5) & $>$500k\,\scriptsize(0/5) & $>$500k\,\scriptsize(0/5) \\
 & 100\% & 60k & \textbf{37.5k} & 265k & $>$500k\,\scriptsize(0/5) & $>$500k\,\scriptsize(0/5) & $>$500k\,\scriptsize(0/5) & $>$500k\,\scriptsize(0/5) \\
\addlinespace
\HardMove{8} & 75\% & 45k & \textbf{40k} & 262k & $>$500k\,\scriptsize(0/5) & $>$500k\,\scriptsize(0/5) & $>$500k\,\scriptsize(0/5) & $>$500k\,\scriptsize(0/5) \\
 & 90\% & 62.5k & \textbf{47.5k} & 275k & $>$500k\,\scriptsize(0/5) & $>$500k\,\scriptsize(0/5) & $>$500k\,\scriptsize(0/5) & $>$500k\,\scriptsize(0/5) \\
 & 100\% & 75k & \textbf{72.5k} & 322k & $>$500k\,\scriptsize(0/5) & $>$500k\,\scriptsize(0/5) & $>$500k\,\scriptsize(0/5) & $>$500k\,\scriptsize(0/5) \\
\bottomrule
\end{tabular}

\caption{Total environment steps to 75\%, 90\% (\cref{tab:efficiency}) and
100\% of \hyar{}'s published final score; same conventions as
\cref{tab:efficiency}, bold among the arms whose five seeds all reach the
threshold. $^\dagger$At its converged, shorter budget. $^\ddagger$With a
hand-coded parameter prior. By this rule the rank of \ours{} is the same at all three thresholds
except on \Platform{} (second at 75\% and 90\%, first at 100\%) and \Goal{}
(first at 75\% and 90\%, second at 100\%, behind \hyar{}).}
\label{tab:threshold}
\end{table}

\subsection{Pareto view per task and on GPU}
\begin{figure}[H]
\centering
\begin{minipage}{0.85\linewidth}\centering
\includegraphics[width=\linewidth]{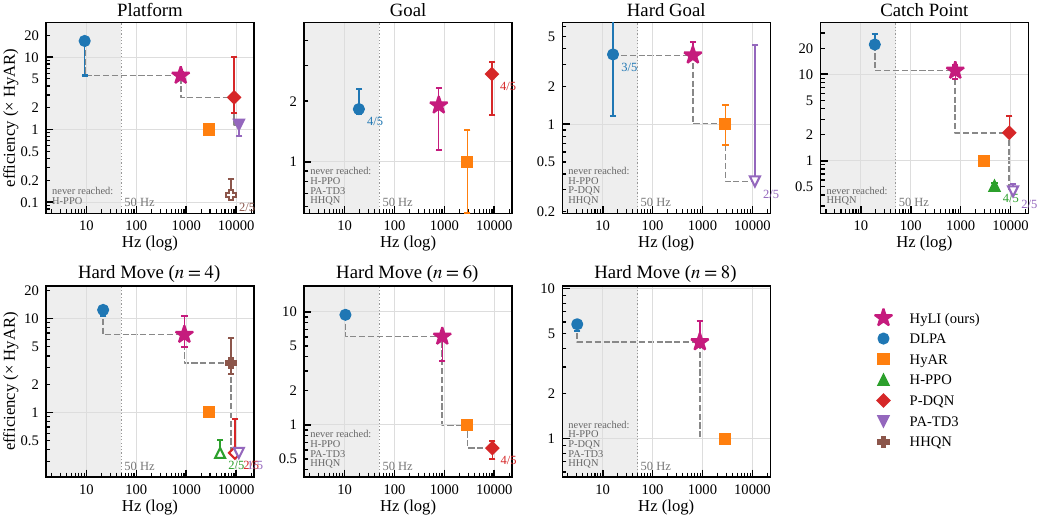}
\end{minipage}
\caption{\Cref{fig:pareto} per task (CPU, one thread). Bars: inter-quartile
range over seeds. Numbers next to a marker: seeds that reach the threshold,
when fewer than five; arms that never reach it are listed in the panel.}
\label{fig:pareto-tasks}
\end{figure}
\begin{figure}[H]
\centering
\begin{minipage}{0.42\linewidth}\centering
\includegraphics[width=\linewidth]{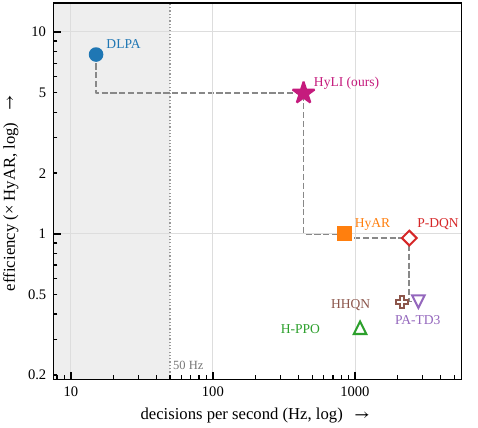}
\end{minipage}\\[4pt]
\includegraphics[width=\linewidth]{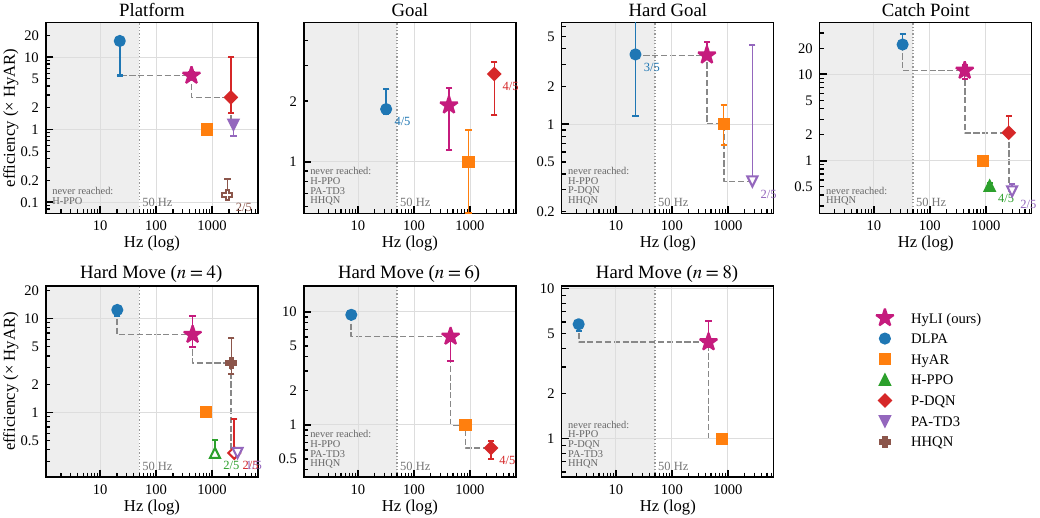}
\caption{\Cref{fig:pareto} with latency measured on the GPU (RTX A4000, batch
size 1). The shape is unchanged.}
\label{fig:pareto-gpu}
\end{figure}

\subsection{Latency on GPU and tail latency}
\begin{table}[H]
\centering\small
\setlength{\tabcolsep}{3.5pt}
\begin{tabular}{lrccccccc}
\toprule
Task & $\Kmodes$ & \oursshort & \dlpa & \hyar & \pdqn & \hhqn & \patdd & \hppo \\
\midrule
\multicolumn{9}{l}{\emph{CPU, 1 thread}} \\
\Platform & 3 & 1.29 (1.31) & 108 (109) & 0.343 (0.346) & 0.11 (0.111) & 0.128 (0.13) & 0.0879 (0.0888) & 0.21 (0.212) \\
\Goal & 3 & 1.3 (1.32) & 51.6 (51.8) & 0.347 (0.349) & 0.112 (0.113) & 0.128 (0.129) & 0.0886 (0.0921) & 0.213 (0.223) \\
H.\,Goal & 11 & 1.56 (1.59) & 62 (62.4) & 0.347 (0.35) & 0.107 (0.108) & 0.128 (0.129) & 0.0891 (0.0929) & 0.212 (0.214) \\
Catch & 2 & 1.28 (1.3) & 52.4 (52.6) & 0.337 (0.341) & 0.106 (0.108) & 0.128 (0.13) & 0.0889 (0.0933) & 0.209 (0.211) \\
\HM{4} & 16 & 1.09 (1.11) & 46.2 (46.5) & 0.345 (0.348) & 0.106 (0.107) & 0.128 (0.129) & 0.0883 (0.0895) & 0.21 (0.212) \\
\HM{6} & 64 & 1.1 (1.12) & 95.9 (98.2) & 0.346 (0.356) & 0.109 (0.11) & 0.16 (0.161) & 0.11 (0.113) & 0.215 (0.218) \\
\HM{8} & 256 & 1.13 (1.15) & 323 (332) & 0.357 (0.363) & 0.124 (0.129) & 0.149 (0.15) & 0.0968 (0.0976) & 0.242 (0.246) \\
\HM{10} & 1024 & 1.29 (1.31) & 1145 (1151) & 0.381 (0.385) & 0.222 (0.224) & 0.249 (0.251) & 0.164 (0.166) & 0.358 (0.36) \\
\midrule
\multicolumn{9}{l}{\emph{GPU (NVIDIA RTX A4000)}} \\
\Platform & 3 & 2.33 (2.38) & 44.4 (45.6) & 1.25 (1.29) & 0.462 (0.481) & 0.53 (0.566) & 0.415 (0.425) & 1.01 (1.69) \\
\Goal & 3 & 2.37 (2.6) & 31.4 (39) & 1.06 (1.17) & 0.368 (0.377) & 0.394 (0.403) & 0.302 (0.312) & 0.886 (0.912) \\
H.\,Goal & 11 & 2.37 (2.68) & 44.7 (45.5) & 1.18 (1.31) & 0.408 (0.427) & 0.473 (0.484) & 0.365 (0.375) & 0.927 (0.964) \\
Catch & 2 & 2.38 (2.71) & 30.6 (30.8) & 1.1 (1.28) & 0.389 (0.398) & 0.438 (0.447) & 0.339 (0.348) & 0.851 (0.875) \\
\HM{4} & 16 & 2.22 (2.58) & 49.2 (49.7) & 1.27 (1.31) & 0.401 (0.409) & 0.452 (0.461) & 0.35 (0.356) & 0.886 (0.909) \\
\HM{6} & 64 & 2.22 (2.55) & 132 (133) & 1.2 (1.32) & 0.42 (0.432) & 0.474 (0.486) & 0.36 (0.37) & 0.91 (0.935) \\
\HM{8} & 256 & 2.22 (2.4) & 462 (464) & 1.25 (1.4) & 0.462 (0.469) & 0.518 (0.527) & 0.382 (0.399) & 0.985 (1.02) \\
\HM{10} & 1024 & 2.21 (2.45) & 1778 (1794) & 1.24 (1.32) & 0.576 (0.796) & 0.562 (0.578) & 0.429 (0.436) & 0.928 (0.945) \\
\bottomrule
\end{tabular}

\caption{Milliseconds per decision, batch size 1, median (p90) over 1\,000
decisions; CPU with one thread and GPU. At batch size 1 the GPU is slower than
one CPU thread for every policy network.}
\label{tab:e6-full}
\end{table}

\subsection{Learning curves: full budgets, log axis and episode return}
\begin{figure}[H]
\centering
\includegraphics[width=\linewidth]{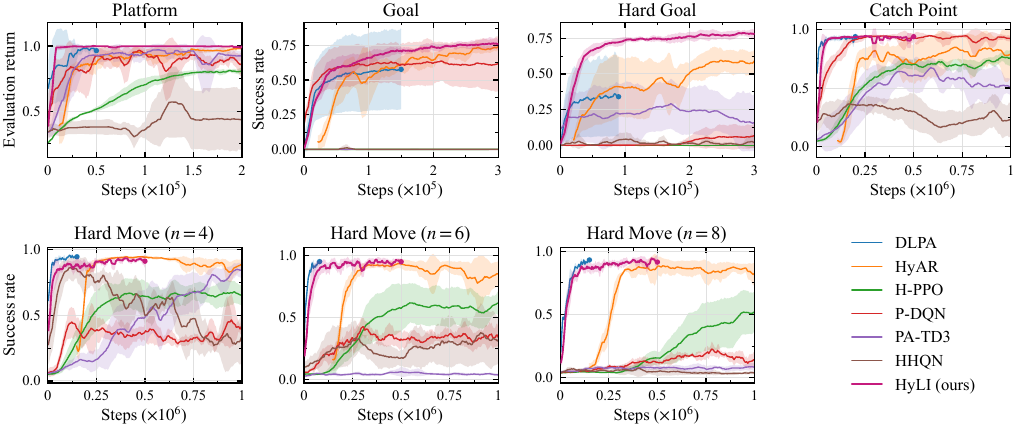}
\caption{\Cref{fig:curves} with every arm drawn over its full budget. As
there, \hyar{}'s reward-free pre-training steps are included on the step axis,
so its curve starts later.}
\label{fig:curves-full}
\end{figure}
\begin{figure}[H]
\centering
\includegraphics[width=\linewidth]{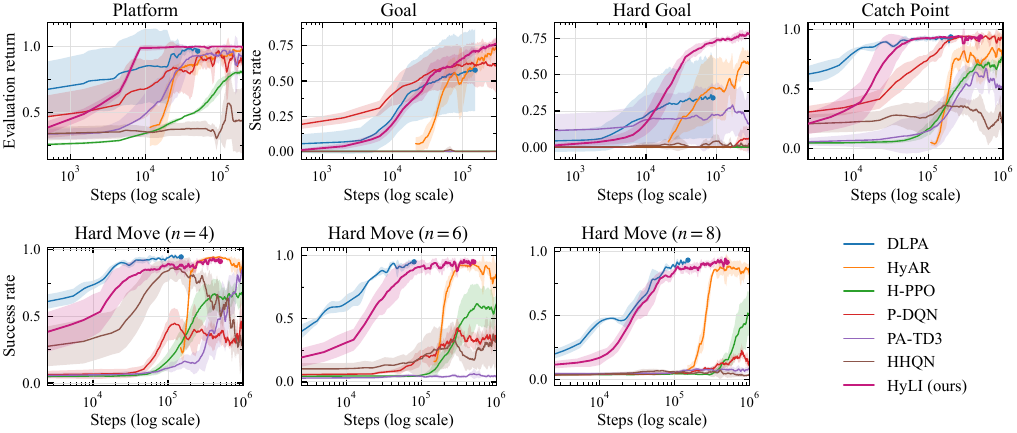}
\caption{\Cref{fig:curves-full} on a logarithmic step axis, which makes the
short \dlpa{} budgets readable. \hyar{}'s pre-training steps are included on
the step axis, as in \cref{fig:curves}.}
\label{fig:curves-log}
\end{figure}
\begin{figure}[H]
\centering
\includegraphics[width=\linewidth]{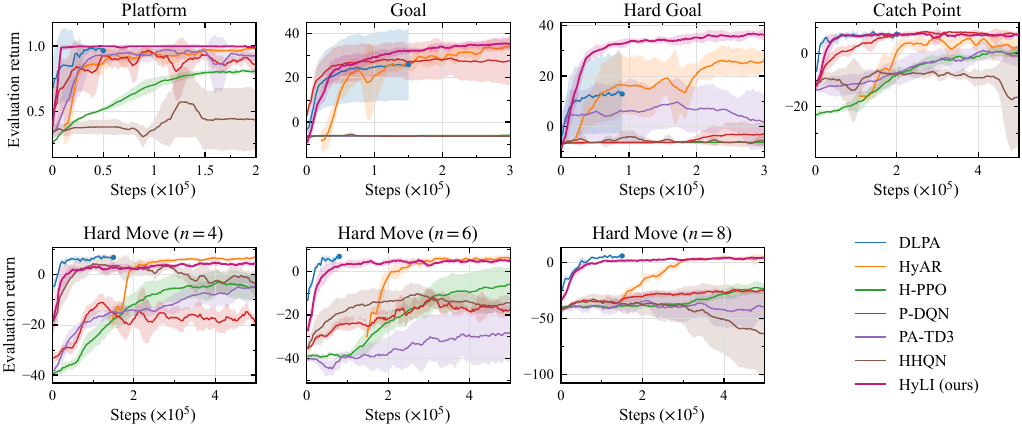}
\caption{\Cref{fig:curves} in evaluation episode return, the quantity plotted
by \citet{li2022hyar} and \citet{dlpa}. On Hard Move, \hyar{}'s return lies
above ours even where its success rate does not, because its successful
episodes are shorter.}
\label{fig:curves-return}
\end{figure}

\subsection{Budgets}
\begin{table}[H]
\centering\small
\setlength{\tabcolsep}{4pt}
\begin{tabular}{lccccccc}
\toprule
Task & \ours{} (ours) & \dlpa$^\dagger$ & \hyar & \pdqn$^\ddagger$ & \hhqn & \patdd & \hppo \\
\midrule
\Platform & 2e5 & 5e4 & 2e5$^\circ$ & 2e5$^\circ$ & 2e5$^\circ$ & 2e5$^\circ$ & 2e5$^\circ$ \\
\Goal & 3e5 & 1.5e5 & 3e5 & 3e5 & 3e5 & 3e5 & 3e5 \\
\HardGoal & 3e5 & 9e4 & 3e5 & 3e5 & 3e5 & 3e5 & 3e5 \\
\CatchPoint & 5e5 & 2e5 & 1e6 & 1e6 & 1e6 & 1e6 & 1e6 \\
\HardMove{4} & 5e5 & 1.5e5 & 1e6 & 1e6 & 1e6 & 1e6 & 1e6 \\
\HardMove{6} & 5e5 & 8e4 & 1e6 & 1e6 & 1e6 & 1e6 & 1e6 \\
\HardMove{8} & 5e5 & 1.5e5 & 1e6 & 1e6 & 1e6 & 1e6 & 1e6 \\
\bottomrule
\end{tabular}

\caption{Budget of every cell in environment steps, as run (\hyar{}'s excludes its
pre-training). $^\circ$: at least one seed's last evaluation lies before 99\% of
the budget.}
\label{tab:budget}
\end{table}

\subsection{Parameter heads}

\begin{table}[H]
\centering\small
\setlength{\tabcolsep}{3pt}
\begin{tabular}{llcccccc}
\toprule
Task & Head & Head width & Score & IQM [95\% CI] & Worst & Fail & Steps to thr. \\
\midrule
Goal & sub$^\star$ & default & \textbf{0.77\,$\pm$\,0.03} & 0.77 [0.74, 0.80] & 0.74 & 0/5 & 45k \\
 & multi & default & 0.73\,$\pm$\,0.03 & 0.72 [0.69, 0.77] & 0.69 & 0/5 & 57.5k \\
 & multi, param-matched & 552 & \textit{0.75\,$\pm$\,0.10} & 0.78 [0.62, 0.83] & 0.57 & 0/5 & 45.5k\,\scriptsize(4/5) \\
\midrule
Hard Goal & sub$^\star$ & default & \textbf{0.79\,$\pm$\,0.04} & 0.80 [0.73, 0.82] & 0.70 & 0/5 & 29.5k \\
 & multi & default & \textit{0.77\,$\pm$\,0.03} & 0.77 [0.74, 0.81] & 0.74 & 0/5 & 27.5k \\
 & multi, param-matched & 1225 & \textit{0.77\,$\pm$\,0.04} & 0.76 [0.72, 0.81] & 0.72 & 0/5 & 19.5k \\
\midrule
Hard Move ($n{=}4$) & sub & default & \textit{0.90\,$\pm$\,0.03} & 0.91 [0.86, 0.93] & 0.85 & 0/5 & 30k \\
 & multi$^\star$ & default & \textbf{0.91\,$\pm$\,0.05} & 0.93 [0.84, 0.95] & 0.81 & 0/5 & 27.5k \\
\bottomrule
\end{tabular}

\caption{Parameter-head comparison of \cref{fig:e2}: final score, mean\,$\pm$\,std
over 5 seeds (bold and italic as in \cref{tab:e1}; $^\star$: the head the method uses),
inter-quartile mean, worst seed, failing seeds (final $<0.5$) and median steps to
90\% of \hyar{}'s published score. Head width: hidden width of the parameter
head (default: the MLP width of \cref{tab:hyper}).}
\label{tab:e2}
\end{table}

\subsection{Ablation learning curves}
\label{app:extra:ablation}
\begin{figure}[H]
\centering
\begin{minipage}{0.48\linewidth}\centering
\includegraphics[width=\linewidth]{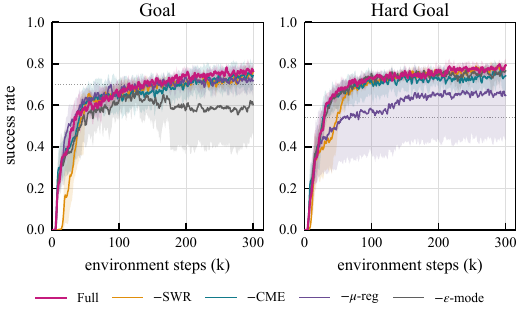}
\end{minipage}
\caption{Leave-one-out ablation of \cref{tab:ablation}: evaluation success rate,
mean over 5 seeds with a 95\% pointwise bootstrap band.}
\label{fig:loo}
\end{figure}


\section{Hardware and compute}
\label{app:compute}
\paragraph{Hardware.} Every run used one of several identical workstations:
one NVIDIA RTX A4000, an Intel x86-64 CPU with 20 logical
cores and 64\,GB of RAM. Each seed ran as its own process with one PyTorch CPU
thread. Latency
(\cref{sec:exp:e6}) was measured on one of these machines, on the CPU with one
thread and on the GPU.

\paragraph{Compute.} Up to five seeds shared one GPU. A seed of ours took
tens of hours; a seed of \dlpa{} took tens to hundreds of hours at budgets 2
to 6$\times$ shorter than ours, because it plans at every training step; the
model-free arms took between one and a few tens of hours. Wall clock depended on how many processes shared a machine, so
we do not compare arms by it; the controlled measurements are the update time
(\cref{tab:cost}) and the decision time (\cref{tab:e6}).

\end{document}